\documentclass[10pt,twocolumn,letterpaper]{article}

\usepackage{wacv}
\usepackage{times}
\usepackage{epsfig}
\usepackage{graphicx}
\usepackage{amsmath}
\usepackage{amssymb}
\usepackage{booktabs}
\usepackage{wrapfig}
\usepackage{cite}
\usepackage{soul}
\usepackage{xcolor}
\usepackage{pifont}
\usepackage{multirow}
\usepackage{algorithm2e}
\usepackage{nicefrac}
\usepackage{comment}
\RestyleAlgo{ruled} 
\usepackage{caption}
\usepackage{subcaption}
\usepackage{microtype}
\newcommand{\algrule}[1][.2pt]{\par\vskip.5\baselineskip\hrule height #1\par\vskip.5\baselineskip}
\usepackage[title]{appendix}

\newcommand{\cmark}{\ding{51}}%
\newcommand{\xmark}{\ding{55}}%

\makeatletter
\DeclareRobustCommand\onedot{\futurelet\@let@token\@onedot}
\def\@onedot{\ifx\@let@token.\else.\null\fi\xspace}

\def\etal{\emph{et al}\onedot}
\makeatother

\def\wacvPaperID{1550} % Enter the WACV Paper ID here

\wacvalgorithmstrack   % Uncomment this line if you are submitting to the Algorithms Track.
\wacvfinalcopy % *** Uncomment this line for the final submission

\ifwacvfinal
\usepackage[breaklinks=true,bookmarks=false,hyperfootnotes=false]{hyperref}
\else
\usepackage[pagebackref=true,breaklinks=true,colorlinks,bookmarks=false,hyperfootnotes=flase]{hyperref}
\fi

\begin{document}
\everypar{\looseness=-1}

%%%%%%%%% TITLE
\title{SITA: Single Image Test-time Adaptation}

% \author{Ansh Khurana\\
% Google Research\\
% \and
% Sujoy Paul\\
% Google Research\\
% \and
% Piyush Rai\\
% IIT Kanpur\\

% }

\author{
Ansh Khurana$^{1*}$, Sujoy Paul$^2$, 
Piyush Rai$^{3*}$, Soma Biswas$^4$, Gaurav Aggarwal$^2$ \\
$^1$Stanford University, $^2$Google Research, $^3$IIT Kanpur, $^4$IISc Bangalore \\
}

\maketitle
\thispagestyle{empty}

%%%%%%%%% ABSTRACT
\begin{abstract}
In Test-time Adaptation (TTA), given a source model, the goal is to adapt it to make better predictions for test instances from a different distribution than the source. Crucially, TTA assumes no access to the source data or even any additional labeled/unlabeled samples from the target distribution to finetune the source model. In this work, we consider TTA in a more pragmatic setting which we refer to as SITA (Single Image Test-time Adaptation). Here, when making a prediction, the model has access only to the given single test instance, rather than a batch of instances, as typically been considered in the literature. This is motivated by the realistic scenarios where inference is needed on-demand instead of delaying for an incoming batch or the inference is happening on an edge device (like mobile phone) where there is no scope for batching. The entire adaptation process in SITA should be extremely fast as it happens at inference time. To address this, we propose a novel approach AugBN that requires only a single forward pass. It can be used on any off-the-shelf trained model to test single instances for both classification and segmentation tasks. AugBN estimates normalization statistics of the unseen test distribution from the given test image using only one forward pass with label-preserving transformations. Since AugBN does not involve any back-propagation, it is significantly faster compared to recent test time adaptation methods. We further extend AugBN to make the algorithm hyperparameter-free. Rigorous experimentation show that our simple algorithm is able to achieve significant performance gains for a variety of datasets, tasks, and network architectures.
\end{abstract}
{\let\thefootnote\relax\footnote{{$^*$Work done while at Google Research.}}}
% \footnote{}
%%%%%%%%% BODY TEXT

\section{Introduction} \label{sec:intro}

\begin{figure}[htbp]
    \centering
    \includegraphics[scale=0.49]{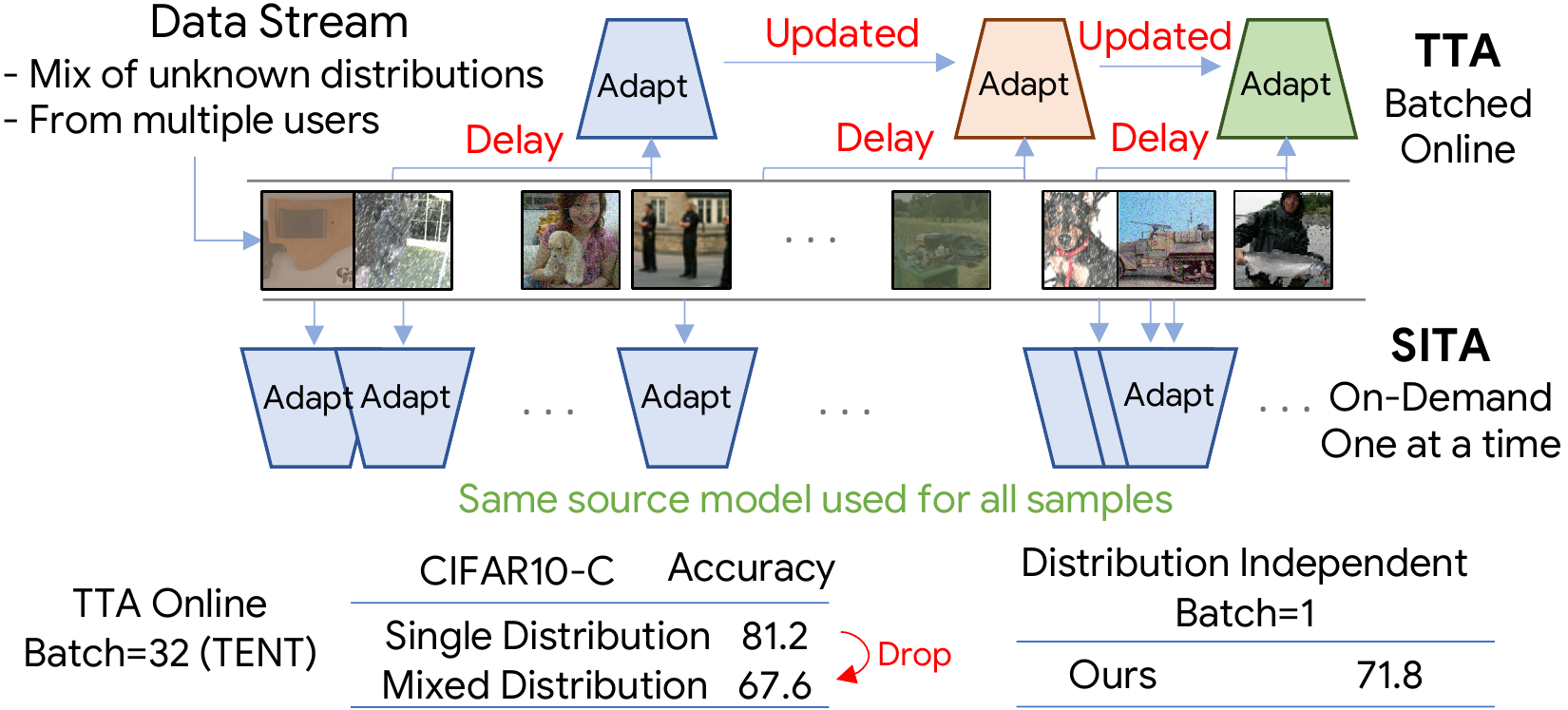}
    \caption{\small \textbf{SITA}: \textbf{S}ingle \textbf{I}mage \textbf{T}est-time \textbf{A}daptation. SITA is a realistic version of test-time adaptation setting. The model only has access to the given test instance, rather than a batch of instances. %SITA is particularly useful where {\em batching} may not be feasible due to privacy and/or latency reasons. 
    Online methods with large batch sizes only work well when encountering test data from a single distribution at a time, and suffer a considerable drop ($81.2$ to $67.6$) in performance when images from different distributions are mixed together and sent as a stream. 
    Contrastingly, in SITA, since adaptation is done for each individual test sample independently, it does not suffer from this problem.}

    \label{fig:motivation}
\vspace{-0.5cm}
\end{figure}

Deep neural networks work remarkably well on a variety of applications, specifically when the test samples are drawn from the same distribution as the training data.  
The performance falls drastically when there is a non-trivial shift between the train and test distributions~\cite{hendrycks2019benchmarking, recht2019imagenet}.
%due to inherent domain shift or corruptions during testing.
%due to natural or artificial reasons. 
In several real-world settings, such a performance drop may make the model unusable. 
There have been recent works in the literature to train robust models \cite{hendrycks2019augmix, sagawa2019distributionally, qiao2020learning}. While this is a viable research direction for the problem at hand, it entails modifying the training process. 
This may not always be practical as the training data may no longer be available due to privacy/storage concerns. 
All that is available is a previously trained model.  
Therefore, there is a growing interest in  \textit{Test Time Adaptation} (TTA), where the models can be adapted at test time, without changing the training process or requiring access to the original training data. 
%This is termed as \textit{Test Time Adaptation} (TTA). 

TTT~\cite{Sun_ICML_2020} and TENT~\cite{Wang_ICLR_2021} are among recent works that have been very effective in adapting models at prediction time.
TTT~\cite{Sun_ICML_2020} uses auxiliary self-supervised tasks to train the source model.
The model is then fine-tuned (via the self-supervised sub-network) for every test instance for multiple iterations. 
Using TTT for adapting a new model, one needs to modify the training process (adding self-supervised sub-network) and hence needs access to the source training data, which may not be always available.
Further, the multiple backward passes take considerable amount of time which may not be available where latency is not acceptable.
%This assumes access to the source data, which may not be available in many scenarios, and the multi-iteration training takes considerable amount of time. 
TENT~\cite{Wang_ICLR_2021}, on the other hand, adapts the given trained model, without accessing the source data.
TENT assumes that the data comes in batches, with batch size usually much greater than one.
%which has multiple issues as will be discussed subsequently. 
It considers an online setting, where the model adapted to the current instances (batch) is used for adapting the subsequent instances (batches), which implies that the model has information about all the test instances seen till a certain point.
%These assumptions make TENT unviable for the proposed SITA setting which is also reflected in our empirical evaluations.
Motivated by the success and limitations of TTT~\cite{Sun_ICML_2020} and TENT~\cite{Wang_ICLR_2021}, we enumerate the following desirable properties of algorithms developed for the realistic, and challenging SITA protocol
\begin{itemize}
    \setlength\itemsep{-0.1em}
    \item Do not require access to the source training data.
    \item Almost as fast as the original model during inference.
    \item Can adapt to a single test instance and does not require batching. %instead of waiting for a batch of data.  
    \item Model adapted to a test instance should not be used on subsequent instances.  
    \item Devoid of hyper parameter tuning at test time.
\end{itemize}

The first property is not only related to privacy/storage issues, but also speed, as any re-training on the source data would make the approach much slower. 
The motivation behind the third property is latency and privacy. For large batch sizes, one has to wait for a certain number of samples, leading to delays, or club samples from multiple users, which may have privacy concerns. 
%The last point is also partially related to privacy as we will be using statistics from the previously seen test instances. More importantly, the model can degrade if it adapts to adversarial examples or huge shifts in data, which can make it unusable for the next test instances. 
The fourth property is motivated by the fact that different test instances/batches may come from very different distributions, which will adversely affect the performance of the model. Figure \ref{fig:motivation} compares the SITA setting with other TTA settings in literature. Further, we cannot expect validation examples to tune hyperparameters at test-time.
%\color{red}
Online methods which use large batch sizes, like TENT~\cite{Wang_ICLR_2021}, improve performance by evaluating one corruption type at a time (single distribution), i.e., resetting the model for the next corruption type. When evaluated by mixing all 15 corruption types in the CIFAR-10-C~\cite{hendrycks2019robustness} dataset (mixed distribution), their performance drops considerably, as shown in Figure~\ref{fig:motivation}.

%{\color{blue}
Apart from TTT and TENT, another recent work BN \cite{Schneider_Neurips_2020} analyzed the power of calibrating batch normalization statistics of a neural network to make the predictions robust against adversarial corruptions. While not particularly designed for the SITA setting, BN can still be applied for it. BN works by replacing the source statistics with a weighted combination of source and target statistics, where the latter in the SITA setting will be estimated from the given single test instance. This has two challenges - first, the single image statistics may not be reliable enough, and second, the calibration weight plays a significant role on the performance, which may vary across test samples. 
In~\cite{Schneider_Neurips_2020}, the parameter is set empirically which is not practical for SITA.
%It is not clear from the work~\cite{Schneider_Neurips_2020} how to choose it.
%}

\begin{table*}[!ht]
\vspace{-0.8cm}
\scriptsize
\caption{Characteristics of various problem settings that deal with adapting a model (trained on a source distribution) to a test distribution. We propose SITA (Single Image Test-time Adaptation), which is the hardest adaptation setting. 
% For each setting we have identified the default characteristics used, which however may be constrained, albeit with a loss in performance. 
%\AK{ Note that the methods proposed in these settings can be adapted to SITA, but are designed to exploit the characteristics of these settings for maximum performance gains.}
%\AK{Mention TTT/FTTA allows for these settings and use them to get better results}}. % \SP{Need to reformat this as it looks very similar to TENT paper.} %
}
\label{table:problem_setting}
\centering
\resizebox{0.9\textwidth}{!}{
\begin{tabular}{lccccccc}
	    \toprule
	    Setting                     & Source Data & Target Label $(y^t)$ & Train Loss                                       & Test Loss          & Offline    & Online Statistics & Batched \\
	    \midrule
	    fine-tuning                 & -           & \cmark  &  $\mathcal{L}(x^t, y^t)$                         & -                  & \cmark     & -                 & \cmark \\ 
	    domain adaptation           & $x^s, y^s$  & \xmark       & $\mathcal{L}(x^s, y^s) + \mathcal{L}(x^t, x^s)$  & -                  & \cmark     & -                 & \cmark \\
	    SFDA & \xmark    & \xmark       &   $\mathcal{L}(x^t)$  & -                  & \cmark     & -                 & \cmark \\
	    TTT~\cite{Sun_ICML_2020}         & $x^s, y^s$  & \xmark       & $\mathcal{L}(x^s, y^s) + \mathcal{L}(x^s)$       & $\mathcal{L}(x^t)$ & \xmark     & \cmark            & \cmark \\
	    FTTA~\cite{Wang_ICLR_2021}  & \xmark      & \xmark       & \xmark                                           & $\mathcal{L}(x^t)$ & \xmark     & \cmark            & \cmark \\
	    \midrule 
	    SITA                & \xmark              & \xmark       & \xmark                                           & \xmark             & \xmark     & \xmark            & \xmark  \\
	    \bottomrule
    \end{tabular}
}
\end{table*}

% \begin{tabular}{lccccccc}
% 	    \toprule
% 	    Setting                     & Source Data & Target Data & Train Loss                                       & Test Loss          & Offline    & Online Statistics & Extended Batch \\
% 	    \midrule
% 	    fine-tuning                 & -           & $x^t, y^t$  &  $\mathcal{L}(x^t, y^t)$                         & -                  & \cmark     & -                 & \cmark \\ 
% 	    domain adaptation           & $x^s, y^s$  & $x^t$       & $\mathcal{L}(x^s, y^s) + \mathcal{L}(x^t, x^s)$  & -                  & \cmark     & -                 & \cmark \\
% 	    source free domain adaptation & \xmark  & $x^t$       &   $\mathcal{L}(x^t)$  & -                  & \cmark     & -                 & \cmark \\
% 	    test-time training          & $x^s, y^s$  & $x^t$       & $\mathcal{L}(x^s, y^s) + \mathcal{L}(x^s)$       & $\mathcal{L}(x^t)$ & \xmark     & \cmark            & \cmark \\
% 	    fully test time adaptation  & \xmark      & $x^t$       & \xmark                                           & $\mathcal{L}(x^t)$ & \xmark     & \cmark            & \cmark \\
% 	    \midrule 
% 	    SITA              & \xmark      & $x^t$       & \xmark                                           & \xmark             & \xmark     & \xmark            & \xmark  \\
% 	    \bottomrule
%     \end{tabular}

In this work, we propose {\bf AugBN}, which overcomes the aforementioned limitations and fulfuls the various desiderata of the \textbf{S}ingle \textbf{I}mage \textbf{T}est-time \textbf{A}daptation (SITA) setting.
%Note that this the most difficult and realistic setting that has been considered in the literature till now. 
The proposed method does not assume any access to the source data, adapts to one test instance at a time, and resets the model to the given source model for adapting to every new test instance. 

Similar to BN~\cite{Schneider_Neurips_2020}, our method calibrates the batch-norm statistics.
However, as estimating them from a single test instance is unreliable (as done in \cite{Schneider_Neurips_2020}), we utilize certain augmentations of the single test instance to obtain a robust estimate of the batch-norm statistics. 
Moreover, we propose an entropy-based approach to estimate the calibration parameter for each test instance instead of treating it like a design choice as done in~\cite{Schneider_Neurips_2020}.
The hyper-parameter free approach we propose shows consistent performance boosts across a variety of datasets for both classification and segmentation.

%to obtain a better estimate of the batch-norm statistics. 
Our adaptation method uses only one forward pass and thus is quite fast (comparable to using the source model directly), unlike TENT~\cite{Wang_ICLR_2021} and TTT~\cite{Sun_ICML_2020}, which require at least one backward pass. 
The main contributions of this work are summarised as follows:
\begin{enumerate}
    \setlength\itemsep{-0.05em}
    \item We formalise the \textbf{S}ingle \textbf{I}mage \textbf{T}est-time \textbf{A}daptation (SITA) setting.
    \item The proposed hyperparameter-free approach performs fast adaptation in both dense and sparse prediction tasks, with only one forward pass. 
    \item We achieve state-of-the-art performance for SITA on both classification and segmentation tasks.
\end{enumerate}

%%%%% RELATED WORKS

\section{Related Work} \label{sec:related}

Table~\ref{table:problem_setting} summarizes the characteristics of various settings which adapt a model trained on a \emph{source} distribution to a \emph{target} distribution. Fine-tuning and domain adaptation are offline methods, that is, they assume access to the entire source and target dataset to adapt and thus are out of scope for {\em test time} settings. We divide the discussion on related works based on the characteristics of the {\em test time} setting proposed by recent works. 
%The categorization is as follows (1) Test-Time Training, (2) Fully Test Time Adaptation, (3) BatchNorm Adaptation. 

{\flushleft \bf{Source-free Domain Adaptation.}} One of the constraints of SITA is that we cannot use the source data while adaptation, but only use the trained source model. Recent works in literature tackle this problem, but using an unlabeled target set for adaptation, with the hypothesis that the test instances are drawn from the target distribution. These methods include - entropy minimization with divergence maximization \cite{liang2020we}, pseudo-labeling with self-reconstruction \cite{yeh2021sofa}, as well as generating additional target images \cite{li2020model, Liu_CVPR_2021} and robustness to dropout \cite{S_2021_CVPR}. Unlike these methods, in SITA, we do not have any target set to adapt and do not have any assumption over the distribution of the test instances. 

{\flushleft \bf{Test Time Training.}} In this setting, self-supervised tasks are introduced during training of the source models. These self-supervised tasks are later used at test time, allowing adaptation of the shared encoder between the main prediction branch and the auxiliary self-supervised task branch. TTT~\cite{Sun_ICML_2020}, one of the first works to formalize test time training, uses the self-supervised task of rotation prediction. %Karani Other recent works~\cite{Bartler2021MT3MT} extend the auxiliary branch based approach by using meta-learning to ensure that test-time training on the auxiliary branch would improve predictions from the main branch. These approaches cannot utilise any off-the-shelf model and improve its performance at test time. Instead they need to re-train the source model with  hand-crafted and often complicated training strategies. 
These approaches have the advantage of adapting off-the-shelf models, since they do not rely on any auxiliary task.
The state-of-the-art method in this category,  TENT~\cite{Wang_ICLR_2021}, formalised this setting and achieved better performance than some of the methods which actually re-train the source model. 
TENT runs an online optimization over a given test set, which minimizes the entropy of the predicted distribution for every incoming batch. Although the method can be made to work in the SITA setting where the model is reset after every test instance, their results are primarily focused in the online setting by gradually adapting the model over the test set. 
It gives impressive performance in the online setting, but we observe significant drop in performance in the SITA setting (Figure~\ref{fig:classification_batchsize}), when there is only a single test instance.
\cite{Mummadi_Arxiv_2021} improves upon TENT by adjusting the loss function to a log-likelihood ratio instead of entropy, and maintaining a running estimate of the output distribution. 
This setting, while clearly more useful than the previous one, is not as realistic as SITA because it needs to (i) accumulate and store samples to create batches, (ii) run an online optimisation under the assumption that the test samples come from the same distribution.

{\flushleft \bf{BatchNorm Adaptation.}} Recent literature on domain-adaptation 
%and improving robustness to corruption 
proposes adapting the statistics of only the Batch Normalization layers for adjusting to the test distribution. 
Though these works are not targeted towards the TTA task, they suggest that adapting normalization statistics could provide significant performance gains, without requiring hand-crafted loss functions.
Prediction Time Normalization (PTN)~\cite{Nado_Arxiv_2020} uses the mean and variance of the current test batch as the statistics in the batch norm layer, instead of using the accumulated statistics of the source data. 
This works reasonably well if the test batch size is large enough to provide a good estimate. 
BN \cite{Schneider_Neurips_2020} uses a similar approach with focus on improving robustness to corruption.
The source model's accumulated statistics are combined with statistics accumulated over all the available test images to reach a reliable estimate
%They show usefulness of the approach 
in two settings: (i) \emph{full adapt} where the entire test set is available and (ii) \emph{partial adapt} where a subset of the test set is available. 
A special case of the \emph{partial adapt} setting, where only one image is available at a time can be considered as test time adaptation. 
Our AugBN algorithm is a significant improvement over BN. AugBN obtains a robust estimate of batch-norm statistics from single image using augmentations.
Additionally, the proposed approach automatically finds an optimal calibration parameter for each single test instance independently- without any validation dataset. These improvements make our method better applicable to the challenging SITA setting, showing consistent performance gains across a variety of datasets and tasks.
There are a few concurrent works related to TTA which recently appeared online. 
%We discuss them here for completeness. 
You~\etal~\cite{You_Arix_2021} use BN~\cite{Schneider_Neurips_2020} with CORE~\cite{Jin_ECCV_2020} loss to adapt the affine parameters of the batch norm layer. Their method needs backpropagation and large batch sizes, not conforming with SITA. Zhang~\etal~\cite{Zhang_Arxiv_2021} use 32/64 augmented samples for each test sample to arrive at a marginal output distribution and optimizes it using entropy loss similar to TENT~\cite{Wang_ICLR_2021}, requiring an expensive optimization per sample. Hu~\etal~\cite{Hu_Arxiv_21} maintain an online estimate for the statistics of the incoming test data along with augmentations. In comparison, our contributions are unique as we propose a lightweight adaptation technique, in accordance with the harder SITA setting. %{\color{red} 
Further, most of these approaches involve hyper-parameter tuning for best results, and the papers do not address tuning them at test-time, thus making them unrealistic in the SITA setting.

%%%%% METHOD
\section{Methodology} \label{sec:method}
We next discuss in detail the SITA setting and propose a simple, yet effective solution for the same. We first formally define the problem statement, and then discuss the challenges that motivate our approach.
{\flushleft \bf{Problem Statement:}} 
%\subsection{Problem Statement}
Consider that we have a model $f_\theta: \boldsymbol{\mathrm{x}} \rightarrow \boldsymbol{\mathrm{y}}$, trained on a dataset, which we refer to as the {\em source}. 
This model may not perform well on test data drawn from distributions other than the source, which can be either a corrupted version of the source itself or images drawn from a different distribution, both of which will be referred to as the {\em target}. 
The goal is to adapt the model $f_\theta$ such that the adapted model performs better on the target images compared to directly using the source model.
As motivated in Section \ref{sec:intro}, we focus on the challenging SITA setting, where inference has to be done on a single image at a time, and not a batch of instances, with the model being reset to the source model after every test instance adaptation. 
%The setting can apply to both sparse and dense prediction tasks.

%\color{magenta} [SB: These can be removed]
%In particular, we focus on episodic version of the setting, i.e., once a model has been adapted for a test image, we do not use the adapted model to infer the subsequent test instances, but rather go back to the original source model $f_\theta$ and adapt it, for every new test instance. 
%This is a special case of SITA, which we believe is realistic and the hardest from adaptation point of view. \color{black}

%The label $\boldsymbol{\mathrm{y}}$ can be a scalar for classification-like tasks or an image for segmentation-like tasks.

{\flushleft \bf{Internal Covariate Shift:}}
%\subsection{Internal Covariate Shift}
In deep neural networks, the mean and covariance statistics of every batch-normalization layer are learned using exponential moving average during training, and these accumulated statistics are usually used during testing, as given below for a particular layer:
%instead of the test batch statistics, as follows for a particular layer:
\begin{equation}
    \mathrm{BN}(\mathrm{F}) = \gamma \frac{\mathrm{F}-\mathbb{E}_s[\mathrm{F}]}{\sqrt{\mathrm{Var}_s[\mathrm{F}]}} + \beta
    \label{eqn:train-bn}
\end{equation}
where $\mathrm{F}$ is the input feature to the batch norm layer and $\gamma$,  $\beta$ are scale and shift parameters, also learned during training.
The subscript $s$ indicates source data. 
The underlying reason behind using the train statistics instead of the test batch statistics is that the train statistics are estimated on a much larger set compared to a batch of test data, which is much smaller and hence can give a biased estimate of the statistics. 
While this method works well for the test samples drawn from a distribution similar to the source, its performance may degrade if the test samples come from a different distribution. This is because the feature distribution $\mathrm{F}$ of the intermediate layers of the network may have mean and variance shifted from the train statistics, due to the shift in the input distribution. This is often termed as internal covariate shift~\cite{Ioffe_ICML_2015, Schneider_Neurips_2020}. % \SP{maybe add a figure demonstrating the difference in features of train vs test}.
% \AK{Should add some references here for internal covariate shift}

To mitigate this problem, the statistics can be estimated from the target dataset and used instead of the learned source statistics, i.e., replace $\mathbb{E}_s[\mathrm{F}]$ and $\mathrm{Var}_s[\mathrm{F}]$ in Eqn. (\ref{eqn:train-bn}) with $\mathbb{E}_t[\mathrm{F}]$ and $\mathrm{Var}_t[\mathrm{F}]$, which can be computed as follows:
\begin{align}
    \mathbb{E}_t[\mathrm{F}] &= \frac{1}{n_tHW}\sum_{i,j,k} \mathrm{F}_{i,j,k} \nonumber \\
    \mathrm{Var}_t[\mathrm{F}] &= \frac{1}{n_tHW}\sum_{i,j,k} (\mathrm{F}_{i,j,k} - \mathbb{E}_t[\mathrm{F}])^2
    \label{eqn:target-stats}
\end{align}
%\color{magenta} i used for two different things, change it. \color{black}
$\mathrm{F}_i \in \mathbb{R}^{H, W, F}$ is the feature map of the $i^{th}$ instance. $H, W, F$ are the height, width and feature dimension respectively. $j \in \{1, \dots, W\}, k \in \{1, \dots, H\}$ and $n_t$ is the number of target instances in the dataset. This is exactly what is done in several domain adaptation works \cite{li2016revisiting, chang2019domain}, %{\color{blue} 
as well as in Prediction Time Normalization (PTN) \cite{Nado_Arxiv_2020}. The domain adaptation works assume access to a target training set with which one can obtain a good estimate of the above mean and variance. On the other hand PTN assumes a huge batch of test instances, thus able to obtain a good estimate of the mean and variance. However, in the SITA setting, neither do we have access to a target training set or a batch of test instances, and cannot use the updated model or the current instance's statistics to infer the subsequent test instances.
%}
This makes our setting much harder compared to other settings in literature on test time adaptation \cite{Wang_ICLR_2021}.

%{\color{blue}
{\flushleft \bf{Batch-Norm Parameter Calibration:}} %As discussed in \cite{Schneider_Neurips_2020}, 
When we have a limited number of target samples from a distribution different from the source, using only the target statistics from Eqn \ref{eqn:target-stats} may not work well, because of unreliable estimates from only a few samples.
Additionally, using only the source statistics may not work well because of internal covariate shift.
%as well as not use only the target statistics of Eqn \ref{eqn:target-stats}, because of unreliable estimate from only a few samples. 
Instead using a weighted combination of the two may be a better alternative:
%}
\begin{align}
    \mu &=  \lambda \mathbb{E}_s[\mathrm{F}] + (1-\lambda) \mu_t \nonumber \\
    \sigma^2 &=  \lambda \mathrm{Var}_s[\mathrm{F}] + (1-\lambda) \sigma^2_t
    \label{eqn:bn}
\end{align}
where $\lambda \in [0, 1]$. This strategy considers the source statistics as a prior. On one hand, $\lambda=1$ results in using the source model directly on the target instances.
In this case, the estimator has high bias. 
On the other hand, $\lambda=0$ results in using only the single test image statistics.
In this case, the variance of the estimator is high. 
The estimator in Eqn.~(\ref{eqn:bn}) provides a balance to bias and variance, with the underlying hypothesis that the target distribution is a shifted version of the source. It can be shown that the mean estimator has $(1-\lambda)^2$ times lower variance and $||(1-\lambda)(\mathbb{E}_s[\mathrm{F}]-\mathbb{E}_t[\mathrm{F}])||$ lower bias magnitude than the worst variance and bias of the two estimators corresponding to $\lambda=1$ and $\lambda=0$.

%{\color{blue}
While this may perform well when the number of target instances is at least a few, in the SITA setting, where we have only one test instance at a time, estimating the target mean and variance from a single instance may not be reliable. % Moreover, it is not quite clear how to set the prior $\lambda$ for every test instance. We next discuss how we solve these problems.
%}

{\flushleft \bf{Augmentation for Statistics Estimation:}}
%\subsection{Augmentation for Statistics Estimation}
%{\color{blue} 
Given a single test instance, we can estimate its statistics ($\mu_t, \sigma_t$) using Eqn.~(\ref{eqn:target-stats}), with $n_t=1$. However, this estimate may be unreliable as it is computed using a single instance. 
%}
Ideally, we want the single image estimate ($\mu_t, \sigma_t$) to be as close as possible to the true statistics of the target distribution. The variance of the above estimators can be reduced by increasing the number of samples. Given that we have only one sample $\boldsymbol{\mathrm{x}}$ at hand, we ask the question - \textit{is it possible to generate more data points to improve the estimates?} While we do not have access to the underlying distribution of the target instances, we can possibly create data points in the vicinity of $\boldsymbol{\mathrm{x}}$. Inspired by the recent works in contrastive learning for representation learning \cite{chen2020simple} and theoretical justifications of pseudo-labeling \cite{wei2020theoretical}, which suggests that using neighborhood samples via augmentation helps in the respective tasks, we adopt this idea for our task to improve the estimate of target image statistics. 

Specifically, we use a set of augmentations to augment $\boldsymbol{\mathrm{x}}$ and obtain $\{\boldsymbol{\hat{\mathrm{x}}_1}, \dots, \boldsymbol{\hat{\mathrm{x}}_n}\}$. This generates features $\{\hat{\mathrm{F}}_1, \dots, \hat{\mathrm{F}}_n\}$ for a certain batch normalization layer, and we use these features along with the original feature $\mathrm{F}$ to obtain a better estimate of the statistics, $\mu_t$ and $\sigma_t$ from the given single image. 
However, as it is hard to control the distribution of the augmented samples, and certain augmentations can outweigh the estimation, instead of assigning the same weight to all the augmented samples as the original sample, we distribute the weight as follows:
\begin{align}
    \mu_t = \mathbb{E}_w(\{\mathrm{F}, \hat{\mathrm{F}}_1, \dots, \hat{\mathrm{F}}_n\}; w=\{\nicefrac{1}{2}, \nicefrac{1}{2n}, \dots, \nicefrac{1}{2n}\}) \nonumber \\
    \sigma_t =\text{Var}_w(\{\mathrm{F}, \hat{\mathrm{F}}_1, \dots, \hat{\mathrm{F}}_n\}; w=\{\nicefrac{1}{2}, \nicefrac{1}{2n}, \dots, \nicefrac{1}{2n}\})
    \label{eqn:aug-stat}
\end{align}

\begin{figure}[htbp]
\vspace{-0.5cm}
    \centering
    \hspace{-2mm}
    \includegraphics[scale=0.32]{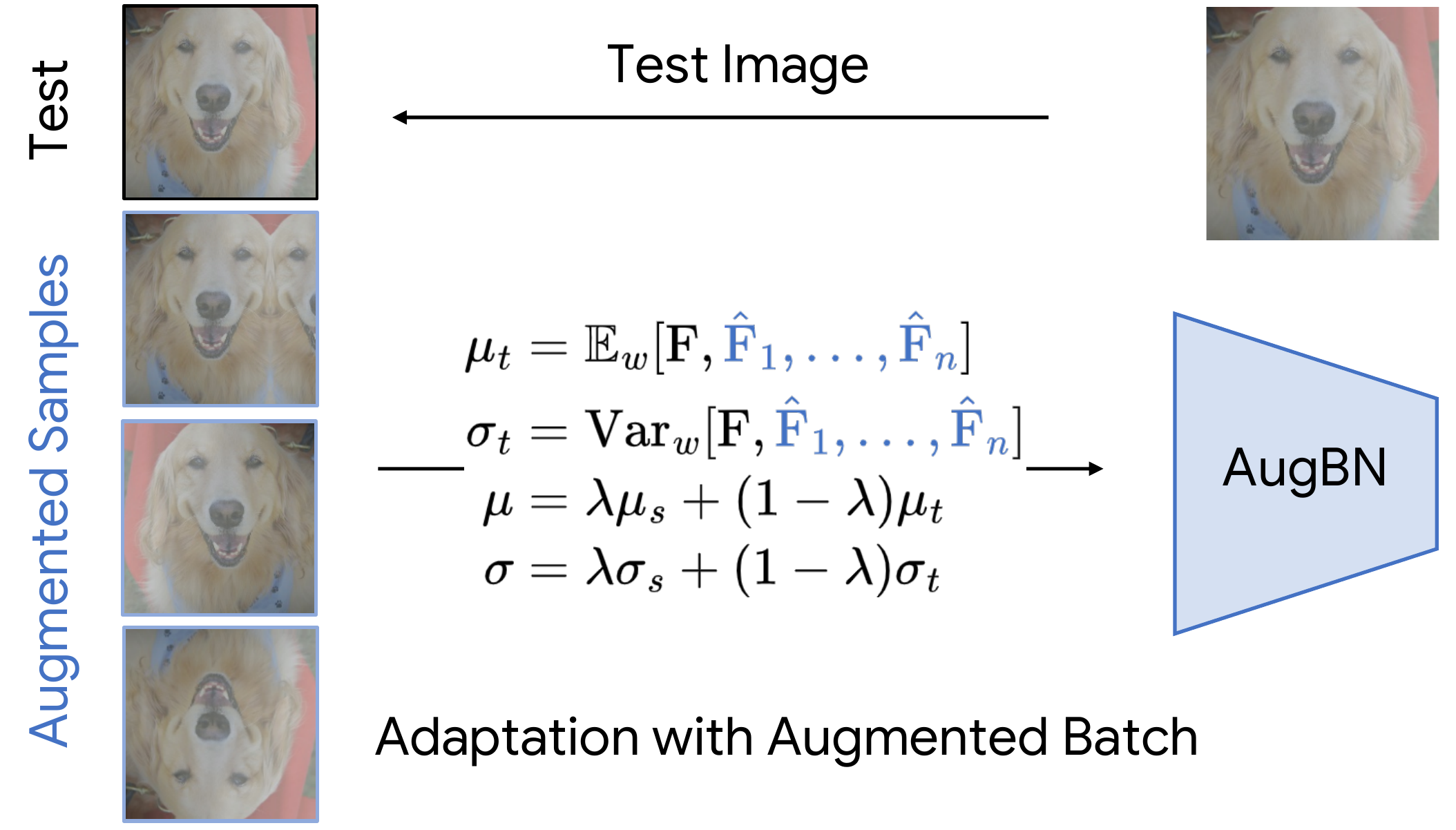}
    \caption{Proposed method (Algorithm ~\ref{main_algo}). The input test image along with a combination of augmented samples are used to estimate the batch-norm statistics using Eqn. \ref{eqn:aug-stat}. This estimation happens with one forward pass through the network using our proposed AugBN layer (Algorithm ~\ref{augbn_algo}).} % \AK{Please review and refine this caption.}} 
    \label{fig:main_algo}
    \vspace{-0.4cm}
\end{figure}
In our work, we choose a fixed set of augmentations $\{a_1, \dots, a_m\}$, from which we randomly pick $k (\leq m)$ augmentations.
These are composed to obtain the augmentation functions $\mathrm{A}_i$, and apply it on the test image to generate augmented images $\boldsymbol{\hat{\mathrm{x}}_i}$.
Since these new samples $\boldsymbol{\hat{\mathrm{x}}_i}$ are generated from a single parent sample $\boldsymbol{\hat{\mathrm{x}}}$, the samples are not independent. Thus, the variance reduction may not be linear with the number of augmented samples used (as the underlying assumption in that case is that the samples are independent). Having said that, using the augmented samples does improve the performance over a range of tasks, as we observe in our experiments (Section~\ref{sec:experiments}).
The entire algorithm to adapt the model during prediction is shown in Algorithm \ref{main_algo}, which uses the proposed AugBN layer (Algorithm \ref{augbn_algo}). A pictorial representation of our proposed method is presented in Figure \ref{fig:main_algo}. 
We must highlight the simplicity of our method, as it can be easily implemented by just replacing the vanilla BN layers in any network with AugBN.

\vspace{-0.2cm}
\begin{algorithm}
\DontPrintSemicolon 
\caption{AugBN}\label{alg:two}
\KwIn{- Source statistics: $\mu_s, \sigma_s, \text{prior}=\lambda$\\ 
$\ \ \ \ \ \ \ \ \ \ \ $ - Input batch features: $\mathrm{F}$, {\color{brown} $\oplus \ \{\hat{\mathrm{F}}\}_{i=1}^n$}.
}
\KwOut{Normalised Features: $\bar{\mathrm{F}} = \mathrm{AugBN}(\mathrm{F}, \{ \hat{\mathrm{F}}_i\}_{i=1}^n)$}
{\color{gray} $\ominus \ \bar{\mathrm{F}} \leftarrow \frac{\mathrm{F} - \mu_s}{\sigma_s} $} \\
{\color{brown} $\oplus \ \mu \leftarrow  \lambda \mu_s  +  (1-\lambda) \mu_t$ (Eqn. \ref{eqn:aug-stat}) \\
$\oplus \ \sigma \leftarrow   \lambda \sigma_s  + (1- \lambda)\sigma_t $ (Eqn. \ref{eqn:aug-stat}) \\
$\oplus \ \bar{\mathrm{F}} \leftarrow \frac{\mathrm{F} - \mu}{\sigma} $ \\}

\algrule
\small{Comment: $\oplus$ and $\ominus$ signify additions and removals from the standard batch normalization layer)}
\label{augbn_algo}
\end{algorithm}

\SetKwComment{Comment}{/* }{ */}
\begin{algorithm}
\DontPrintSemicolon 
\caption{Proposed Algorithm for SITA}\label{alg:one}
\KwIn{- Single Image: $\boldsymbol{\mathrm{x}}$  \\
$\ \ \ \ \ \ \ \ \ \ \ $ - Source Model: $f_\theta$
}
\KwOut{Prediction: $\boldsymbol{\mathrm{y}}$}
$\hat{f}_\theta \leftarrow$ Repalce BN Layer in $f_\theta$ with AugBN Layer  \;
\For{$i=1\dots n$}{
$\mathrm{A}_i \leftarrow \text{Compose} (\text{RAND-Choose-k}(\{a_i\}_{i=1}^m))$ \;
$\boldsymbol{\hat{\mathrm{x}}}_i \leftarrow \mathrm{A}_i(\boldsymbol{\mathrm{x}})$
}
$\boldsymbol{y} \leftarrow \hat{f}_\theta(\boldsymbol{\mathrm{x}}, \boldsymbol{\hat{\mathrm{x}}}_1, \dots, \boldsymbol{\hat{\mathrm{x}}}_n)$
\algrule
\small{Comment: RAND-Choose-k() uniformly randomly chooses $k(<N)$ augmentations from the given set.}
\label{main_algo}
\end{algorithm}
%\AK{3.5 for entropy search for finding lambda automatically.}

% \begin{wrapfigure}{r}{0.35\textwidth}
% \vspace{-2cm}
\begin{figure}
    \captionsetup{font=small}
     \centering
         \includegraphics[scale=0.33]{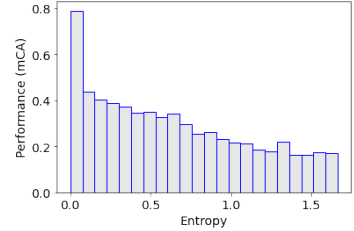}
     \caption{Variation of model performance with entropy.} 
     %For each entropy bin, we plot the percentage of correct predictions among samples whose prediction entropy falls within that bin.}}
     \label{fig:perf_ent_Cifar10}
% \end{wrapfigure}
\end{figure}

%{\color{blue}
{\flushleft \bf{Optimal Prior Selection (OPS):}} We empirically show in Section \ref{sec:experiments} that using the prior ($\lambda$) within a certain range works well in practice. However, it would be even better if we can automatically select the prior for every test instance individually. That is a quite difficult task, as we do not have any information about the test distribution in the SITA setting. To this end, we envisage the probability mass function of the prediction as a metric to measure its correctness. As can be seen from Fig. \ref{fig:perf_ent_Cifar10}, there is strong positive correlation between correctness of prediction and entropy. Specifically, for classification, we run AugBN for $n_p$ different prior values (batched together in a single forward pass)
%between $[0.5, 1.0]$ 
and first choose the top-$k \ (k<n_p)$ priors which have the lowest entropy. We then take a majority voting of the predictions for these top-$k$ priors. For segmentation, we do the same thing to obtain $n_p$ distinct prediction maps for $n_p$ different priors, and then repeat this strategy for all pixels individually. Note that we repeat the same set of augmented images for all the $n_p$ priors. For all datasets, we use $n_p=8, k=3$. 

%%%% EXPERIMENTS

\section{Experiments} \label{sec:experiments}

To demonstrate the effectiveness and universality of our approach, we perform thorough experimentation on a variety of publicly available adaptation datasets for both segmentation and classification tasks.
In comparison, the focus of most related works has primarily been on classification tasks.

\begin{table*}[htbp]
            \small
			\caption{Results for single image test-time adaptation for semantic segmentation and classification. OPS stands for Optimal Prior Selection. % $^*$MixNorm~\cite{Hu_Arxiv_21} and MEMO~\cite{Zhang_Arxiv_2021} are unpublished works on Arxiv and their performance is reported as is. MixNorm~\cite{Hu_Arxiv_21} does not report their source model performance.
    	    }
			\label{table:sota_combined}
			\centering
			\renewcommand{\arraystretch}{1.1}
	        \setlength{\tabcolsep}{4pt}
			\begin{tabular}{l|ccc|cccc}
		    \toprule
		    {} & \multicolumn{3}{c|}{Segmentation (mIoU)} &
		    \multicolumn{4}{c}{Classification (mCA)} \\
		    \midrule
		    \multirow{2}{*}{Methods} & GTA5 & SYNTHIA  & SceneNet & \multirow{2}{*}{CIFAR10-C} & \multirow{2}{*}{ImageNet-C} & \multirow{2}{*}{ImageNet-A} & \multirow{2}{*}{ImageNet-R} \\
		    & $\rightarrow$ Cityscapes & $\rightarrow$ Cityscapes & $\rightarrow$ SUN\\
		  %  \midrule
		  %  Source (MixNorm)  & - & - & - & - & - & - & - \\ 
		    
		    \midrule
		  %  Source (MEMO) & - & - & - & 67.3 & 18.0 & 0 & 36.1 \\ 
		  %  MixNorm$^*$ \cite{Hu_Arxiv_21} & - & - & - & 68.0 & 21.1 & - & - \\ 
		  %  MEMO$^*$ \cite{Zhang_Arxiv_2021} & - & - & - & 70.3 & 24.2 & 0.9 & 41.2 \\ 
		  %  \midrule
		    % \multicolumn{4}{c}{0 Iteration} \\
		    Source Model & 37.6 & 32.1 & 26.5 & 61.4 & 20.5 & 0.5 & 35.4\\
		    {Aug Ensemble} & 35.5 & 32.4 & 26.7 & 61.3& 20.5 & 0.6 & 35.2 \\
		    PTN \cite{Nado_Arxiv_2020, Schneider_Neurips_2020} & 36.7 & 31.8 & 25.2 & 55.8 & 0.3 & 0 & 1.4 \\
		    BN \cite{Schneider_Neurips_2020} & 40.4 & 32.9 & 28.2 & 65.1 & 24.5 & 0.9 & 38.4\\
		  %  BN \cite{Schneider_Neurips_2020} (best)        & 42.3 & 33.7 & 28.3\\
		    TENT \cite{Wang_ICLR_2021} & 37.6 & 32.7 & 26.6 & 55.9 & 0.3 & 0 & 1.2 \\
		    \midrule
		    AugBN & 42.8 & 35.3 & 28.7 & 71.8 & 25.0 & 1.1 & 39.5 \\
		    AugBN+OPS &  \textbf{42.9} &  \textbf{36.5} & \textbf{28.8} & \textbf{73.1} & \textbf{25.5} & \textbf{1.1} & \textbf{40.3} \\
		    \bottomrule
	    \end{tabular}
		%
% 		\vspace{-15pt}
\end{table*}

\subsection{Datasets and Implementation Details} \label{sec:dataset_implement}

{\flushleft \bf{Semantic Segmentation:}} We evaluate our method on three different source $\rightarrow$ target combinations covering not only outdoor scenes, but also indoor scenes, to showcase the wide usability of our algorithm. For outdoor, we evaluate on GTA5 \cite{Richter_ECCV_2016} $\rightarrow$ Cityscapes \cite{cityscapes} and SYNTHIA \cite{Ros_CVPR_2016} $\rightarrow$ Cityscapes. For indoor, we show results on SceneNet \cite{mccormac2017scenenet} $\rightarrow$ SUN \cite{song2015sun}.  The outdoor  and indoor scene datasets have $19$ and $13$ categories, respectively. Refer to Appendix~\ref{app:datasets} for more details. 
%, and also on the indoor setting, which has not been considered before in literature. 
%No previous works have studied test time adaptation on segmentation, except a secondary comment in the results of TENT~\cite{Wang_ICLR_2021}.
%where they mention TENT can scale for dense prediction tasks by running an experiment on the GTA5 \cite{Richter_ECCV_2016} $\rightarrow$ Cityscapes \cite{cityscapes} setting. 
%{\flushleft \bf{Implementation Details.}} 
Following the literature on domain adaptation of semantic segmentation models, we use Deeplab-V2 \cite{deeplab} with ResNet-101 \cite{He_CVPR_2016} as the backbone. We use one GPU to train source models with a batch size of $1$ in all experiments. We use SGD with an initial learning rate of $2.5\times 10^{-4}$ with polynomial decay of power $0.9$~\cite{deeplab}. We use the standard metric of mean intersection over union (mIoU)~\cite{deeplab}. Note that we use only one augmented image and prior ($\lambda=0.8$) in all experiments on segmentation, %{\color{blue}
stated as AugBN. AugBN+OPS however automatically chooses the prior for each test instance independently.
%} 
We use Gaussian blur and random rotation as augmentations, i.e., with $k=m=2$ with number of augments $n=1$, unless otherwise mentioned. (Algorithm \ref{main_algo}). 

{\flushleft \bf{Classification:}} We evaluate the AugBN method for the classification task on common adaptation datasets~\cite{Wang_ICLR_2021, Schneider_Neurips_2020, Mummadi_Arxiv_2021, Sun_ICML_2020}, namely the corruption and perturbation data, CIFAR-10-C and ImageNet-C~\cite{hendrycks2019robustness}. Further, we test our approach on ImageNet-R and ImageNet-A datasets, comprising of renditions of ImageNet classes and adversarial ImageNet examples respectively. Following previous works, we report results on the highest severity level of corruption. For more details of the datasets, please refer to Appendix~\ref{app:datasets}. We compare the mean Classification Accuracy (mCA) over all 15 corruption types on both the datasets. We adopt the same network architecture used in the recent works ~\cite{Wang_ICLR_2021, Sun_ICML_2020}, i.e., ResNet-26 and ResNet-50 for CIFAR-10 and ImageNet respectively. The source model for CIFAR-10 is trained on one GPU with a batch size of 64, using SGD with cosine decay scheduler and an initial learning rate of 0.01. The source model achieves 93.84\% accuracy on the CIFAR-10 test set. For ImageNet, the model is trained on 8x8 TPU slices with a batch size of 8192. We use Adam optimizer along with the cosine decay scheduler with an initial learning rate of 0.1. This source model obtains 76.4\% accuracy on the ImageNet validation set. For ImageNet-C and CIFAR-10-C, we report results with prior $\lambda = 0.9$ and $\lambda = 0.7$ respectively for AugBN. AugBN+OPS automatically chooses the prior for each test image independently.
%} 
For all classification results we use two augmented samples, each composed of five augmentations from color distortion ~\cite{Sun_ICML_2020}, rotation, mirror reflection, vertical and horizontal flip augmentations which are randomly shuffled, i.e., with $n=2, k=5, m=5$ (Algorithm \ref{main_algo}). For more details, please refer to Appendix~\ref{app:augmentations}.

{\flushleft \bf{Baselines.}} 
We compare with the strong state-of-the-art baselines, namely TENT \cite{Wang_ICLR_2021}, BN \cite{Schneider_Neurips_2020}, and Prediction Time Normalization (PTN) \cite{Nado_Arxiv_2020}, and Aug-Ensemble. In Aug-Ensemble, we follow a common practice and augment the test instance to obtain multiple augmented images, and then take an average over all the predictions to get the final prediction. For classification we use the same augments as we use in AugBN. But for segmentation, to maintain the spatial correspondence, we only use gaussian smoothing and gaussian noise as the augmentations. All methods are run in the SITA setting, that is, the test batch size for all experiments is fixed to $1$ and no method is allowed to run an online optimization or to maintain online statistics. Note that the results in the TENT paper~\cite{Wang_ICLR_2021} are for the online setting over a batch size of 64 and 128 for ImageNet-C and CIFAR-10-C respectively. But we apply it in the SITA setting, where we reset the model to the source model after inferring every test instance, and use five iterations of optimization with a learning rate of $10^{-3}$ (this is the best result we obtained by experimenting over a range between $[10^{-2}, 10^{-5}]$) with Adam optimizer as it provides the best computation time vs performance trade-off. For BN, we showcase the results with the suggested hyper-parameter setting ($N=16$ in \cite{Schneider_Neurips_2020}) for the single image case.

\begin{figure*}[htbp!]
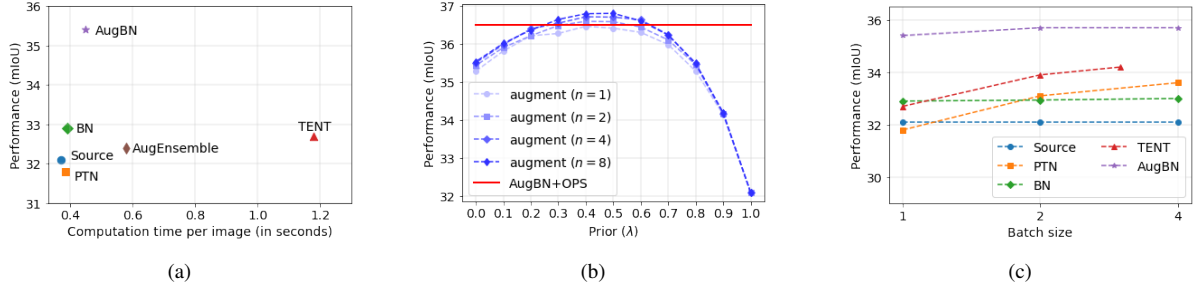

    \captionsetup{font=small}
     \centering
     \begin{subfigure}[b]{0.3\textwidth}
         \centering
         \includegraphics[scale=0.34]{figures/time_comparison_synt2city.pdf}
         \caption{}
         \label{fig:segmentation_time}
     \end{subfigure}
     \begin{subfigure}[b]{0.32\textwidth}
         \centering
         \includegraphics[scale=0.34]{figures/prior_vs_augment_synt2city.pdf}
         \caption{}
         \label{fig:segmentation_prior}
     \end{subfigure}
     \begin{subfigure}[b]{0.32\textwidth}
         \centering
         \includegraphics[scale=0.34]{figures/segmentation_batchsize_synt2city.pdf}
         \caption{}
         \label{fig:segmentation_batchsize}
     \end{subfigure}
     \caption{\small (a) Computation time vs Performance comparison (b) Performance of our algorithm for various augmentations and prior settings  (c) Performance comparison for different batch sizes. TENT cannot fit batch size $>3$ due to the high memory requirement of backprop. All these analyses are performed on SYNTHIA $\rightarrow$ Cityscapes.}
\end{figure*}

\subsection{Comparative Analysis}
{\flushleft \bf{Comparison with state-of-the-art.}}
%{\color{blue}
Table~\ref{table:sota_combined} shows the comparison with state-of-the-art methods for both classification and segmentation. For segmentation, AugBN achieves the best performance on all the three datasets with a 14.8\%, 10.0\% and 7.7\% relative improvement in mIoU over the source model on the GTA5 \cite{Richter_ECCV_2016} $\rightarrow$ Cityscapes \cite{cityscapes} and SYNTHIA \cite{Ros_CVPR_2016} $\rightarrow$ Cityscapes and \cite{mccormac2017scenenet} $\rightarrow$ SUN \cite{song2015sun} respectively. For classification, AugBN gives a huge 17\% relative improvement on CIFAR-10-C compared to just using the source model directly, and also compares favourably with the state-of-the-art approaches. ImageNet-C is a much harder dataset to do adaptation using just a single instance, and the methods which do not incorporate source statistics (TENT~\cite{Wang_ICLR_2021} and PTN~\cite{Nado_Arxiv_2020}) suffer a huge loss in performance as their normalisation statistics are highly incompatible with the source model. Meanwhile, BN~\cite{Schneider_Neurips_2020} and AugBN which add a prior weight towards source statistics can handle adaptation using a single image much better. % \AK{We also report MEMO~\cite{Zhang_Arxiv_2021} and MixNorm~\cite{Hu_Arxiv_21} results from their respective papers for completeness, however these cannot be directly compared as they are obtained from different source models.}

Note that AugBN and BN use a fixed prior value for $\lambda$ as described above in Section \ref{sec:dataset_implement}. This is often set based on what value works well on the validation set if available.
On the other hand, the proposed entropy-based OPS strategy is meant to avoid this specific dependence on the validation set to be able to choose appropriate value of $\lambda$ for each test instance independently. This is a much harder evaluation protocol as it does not use a validation set or have a human-in-the-loop to somehow set an appropriate value.
This hyper-parameter free approach achieves the best performacne across tasks and datasets as shown in Table~\ref{table:sota_combined}. % (bottom section).

%is able to achieve at least the fixed prior performance, and better in some cases, without even having to choose the prior value beforehand. This demonstrates how a light weight statistics update, which is applicable in the challenging SITA setting, could lead to a significant improvement in performance. 
%}

%, with a minimal increase in inference time (Figure~\ref{fig:segmentation_overview}). 

% \input{tables/segmentation_sota}

% \begin{figure}[htbp]
%     \captionsetup{font=small}
%     \centering
%     \includegraphics[scale=0.55]{figures/segmentation_prior.pdf}
%     \caption{Performance of AugBN (relative to the maximum performance obtained) with varying $\lambda$. We used $\lambda=0.8$ for all the segmentation experiments.
%     %This plot presents the performance of our algorithm on the three datasets. For each dataset, the results are relative to the maximum performance obtained for that dataset. 
%     }
%     \label{fig:segmentation_momentum}
% \end{figure}

\begin{figure*}[htbp!]
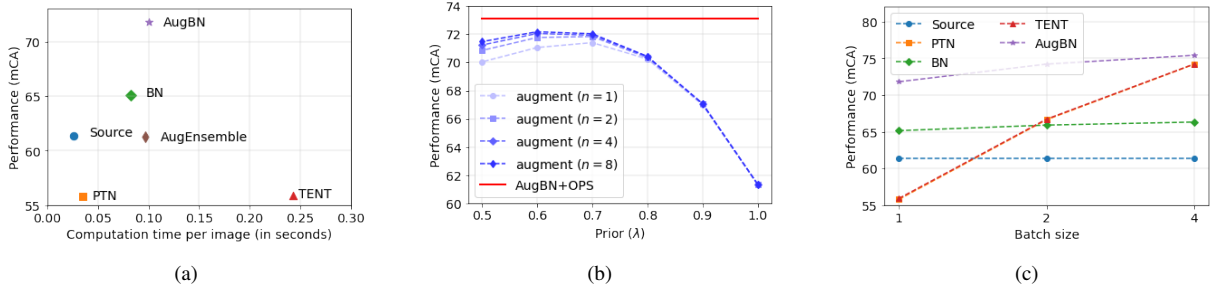

    \captionsetup{font=small}
         \centering
     \begin{subfigure}[b]{0.3\textwidth}
         \centering
         \includegraphics[scale=0.34]{figures/time_comparison_cifar10_with_ensemble.pdf}
         \caption{}
         \label{fig:classification_time}
     \end{subfigure}
     \begin{subfigure}[b]{0.32\textwidth}
         \centering
         \includegraphics[scale=0.34]{figures/prior_vs_augment_cifar10_redraw.pdf}
         \caption{}
         \label{fig:classification_prior}
     \end{subfigure}
     \begin{subfigure}[b]{0.32\textwidth}
         \centering
         \includegraphics[scale=0.34]{figures/classification_batchsize_4_redraw.pdf}
         \caption{}
         \label{fig:classification_batchsize}
     \end{subfigure}
     \vspace{-3mm}
     \caption{(a) Computation time vs performance comparison. (b) Performance of our algorithm for various augmentations and  prior settings. (c) Performance comparison for different batch sizes. All these analysis are performed on CIFAR10-C dataset.}
\end{figure*}

{\flushleft \bf{Computational Analysis}}
%{\color{blue}
The proposed method requires only one forward pass of the original test image itself along with its augmented versions, and thus the inference time is very close to the source model itself. We report the average computation time for all methods on the  SYNTHIA \cite{Ros_CVPR_2016} $\rightarrow$ Cityscapes setting in Figure~\ref{fig:segmentation_time} and on CIFAR-10-C in Figure~\ref{fig:classification_time}. 
This computation time also includes the time required to compute the necessary augmentations.  
Compared to TENT~\cite{Wang_ICLR_2021} which is currently the state-of-the-art TTA method, our method is 2.6x faster with a 10\% relative improvement on segmentation and  (Table~\ref{table:sota_combined}) and 2.5x faster with a 28\% improvement in classification . % The speed of our method can be further improved by embedded implementations for better usage of GPUs. The simplicity of inference in the proposed method and its performance shows how universally applicable it is for any deployed application, as it conforms to the SITA setting. 
%}

{\flushleft \bf{Ablation of design choices.}} 
%{\color{blue}
AugBN uses a combination of augmented samples around the current test image, and a prior factor $\lambda$ to combine the source model's statistics to estimate the target statistics. 
Here, we analyse how the model performance varies with the choice of these parameters. Figure~\ref{fig:segmentation_prior} and Figure \ref{fig:classification_prior} shows the effect of the prior. $\lambda=1$ uses only the source statistics, and hence is equivalent to the source model. $\lambda=0$ on the other hand, ignores source statistics completely, and uses an aggregate of statistics from the single test image and its augmented samples. 
We observe that AugBN is strictly better than using the source model $\lambda=1$.
Further, the results are not very sensitive to variations in $\lambda$, as 20\% of the entire range of possible values gave results within $1$ mIoU of the best result for all the three datasets. Our OPS approach automatically selects $\lambda$ - shown using a horizontal straight line in Figure \ref{fig:segmentation_prior} and \ref{fig:classification_prior}. 

Figure \ref{fig:segmentation_prior} and \ref{fig:classification_prior} also show a combination of changing prior $\lambda$ and the number of augmented images on SYNTHIA $\rightarrow$ Cityscapes. We can see with a higher number of augmented samples, the performance improves as it provides a more reliable estimate. Further, we observe a better stability over the prior $\lambda$ with higher number of augments. 

Though the focus of this work is on SITA setting where only a single test image is available for adaptation, AugBN can also be used for larger batch sizes.
Figure \ref{fig:segmentation_batchsize} and ~\ref{fig:classification_batchsize} shows how performance of all the methods varies with different batch sizes on the SYNTHIA $\rightarrow$ Cityscapes dataset and CIFAR-10-C respectively. 
We observe that AugBN outperforms the other methods even for larger batch sizes. Since TENT requires backward propagation as well, it has a high memory requirement, and thus is not able to fit batch sizes more than $3$ for segmentation, given the high dimensional inputs of size $512 \times 1024$.  As expected, the performance of TENT improves consistently with larger batches.

\begin{figure}[htbp]
    \captionsetup{font=small}
     \centering
         \includegraphics[scale=0.4]{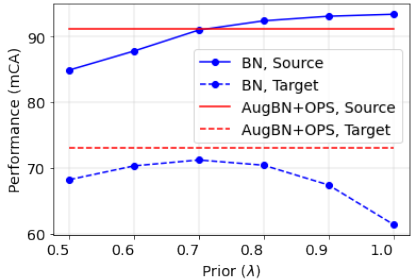}
     \caption{
     %{\color{red} 
     \textbf{Generalized TTA}: The $\lambda$ parameter in BN~\cite{Schneider_Neurips_2020} that works well for images from source distribution (CIFAR10) is expectedly different from the one that works well for the target distribution (CIFAR-10-C) which is a sticky issue. On the other hand, the proposed AugBN+OPS method (horizontal lines) does not need a curated $\lambda$ value, instead it  automatically chooses the best $\lambda$ for each test instance.}
     \label{fig:in_out_cifa10}
 \end{figure}

{\flushleft \bf{Results on additional architectures: }}
To showcase the universality of our approach we test our adaptation technique on across various pretrained ImageNet networks on the ImageNet-C dataset in Table~\ref{table:add_arch_imagenetc_mini}. We provide results for ImageNet-A and ImageNet-R in Appendix~\ref{app:architectures}. Our approach provides significant performance gains across all architectures and datasets.

\begin{table}[htbp!]
            % \vspace{-0.3cm}
			\caption{ImageNet-C results on additional architectures.
    	    }
			\label{table:add_arch_imagenetc_mini}
			\centering
			\renewcommand{\arraystretch}{1.2}
	        \setlength{\tabcolsep}{4pt}
			\begin{tabular}{lcc}
		    \toprule
		    Methods & DenseNet121 & InceptionV3 \\
		    \midrule
		    % \multicolumn{4}{c}{0 Iteration} \\
		    Source & 22.7 & 30.4 \\
		    BN & 25.6 & 35.8\\
		    AugBN & 25.7 & 36.0 \\
		    AugBN+OPS & \textbf{25.8} & \textbf{36.2}\\
		    \bottomrule
	    \end{tabular}
\end{table}

{\flushleft \bf{Generalized Test Time Adaptation:}}
%{\color{blue}
In practice, the test instances can originate from both the source as well as unknown target distributions. Ideally TTA algorithms should work well on both the cases. In the SITA setting, we have no information about the originating distribution, and thus the BN method~\cite{Schneider_Neurips_2020}, which calibrates the batch-norm statistics using a fixed prior, may not work well if a test instance from the source appears (Figure \ref{fig:in_out_cifa10}). This is because the source statistics would work the best in such a case, rather than the calibrated statistics. However, the proposed entropy-based OPS method would automatically choose a suitable calibration parameter $\lambda$ irrespective of whether the test instance is from unknown target distribution or the source distribution. 
%Hence there is no performance degradation on the source samples 
Hence the proposed AugBN+OPS approach is more robust in such realistic situations. 

\section{Conclusion} 
\label{sec:conclusion}
Test-time adaptation has gained recent interest in the literature given its ability to adapt models only during test-time. In this work, we formalise the test time problem under the realistic and challenging \textbf{S}ingle \textbf{I}mage \textbf{T}est-time \textbf{A}daptation (SITA) setting. We propose a simple framework which significantly outperforms the state-of-the-art while overcoming some of the limitations of the existing approaches. 
The strength of the proposed approach is in its simplicity and effectiveness for both semantic segmentation and classification applications. 

\clearpage

%%%%%%%%% REFERENCES
{\small
\bibliographystyle{ieee_fullname}
\bibliography{egbib}
}
\clearpage

%%%% APPENDIX

\begin{appendices}

\section{Datasets}
\label{app:datasets}
\subsection{Semantic Segmentation}

{\flushleft \bf{GTA5 $\rightarrow$ Cityscapes:}} In this setting, we consider GTA5 \cite{Richter_ECCV_2016} as the source and Cityscapes \cite{cityscapes} as the target dataset. The source images are at $760 \times 1280$ resolution, while the target images are used at $512 \times 1024$. The datasets have $19$ categories. The source dataset (GTA5) has $24966$ training images. We perform test time adaptation on the validation set which has $500$ images of Cityscapes. 

{\flushleft \bf{SYNTHIA $\rightarrow$ Cityscapes:}} In this setting, we consider SYNTHIA \cite{Ros_CVPR_2016} as the souce and Cityscapes \cite{cityscapes} as the target dataset. SYNTHIA contains $9400$ training images. However, unlike GTA5, due to lack of proper annotations for a few categories in SYNTHIA, we remove them from evaluation and report the results for $16$ categories, following the literature. The validation set used for test-time adapttaion is again the $500$ images commonly used for Cityscapes evaluation. 

{\flushleft \bf{SceneNet $\rightarrow$ SUN:}} Both of the above two settings are for outdoor scenes, and in this setting we consider indoor scenes with SceneNet \cite{mccormac2016scenenet} as the source and SUN \cite{song2015sun} as the target. The SceneNet dataset has around $5$ million simulated images. However, a lot of the images are rather simple with only a few categories in them. Thus, to train the source model, we only choose the top $50,000$ images having the highest number of categories of SceneNet. For test-time adaptation, we perform experiments on the test set of SUN containing $5050$ images. Both of these datasets contain $13$ categories. Specifically, we use the label transformation available with the SceneNet dataset \footnote{https://github.com/ankurhanda/sunrgbd-meta-data} to map the labels in the SUN dataset such that it matches with the label space of SceneNet. 

\subsection{Classification}
\begin{figure*}[htbp]
    % \hspace{-3mm}
    \centering
    \includegraphics[scale=0.4]{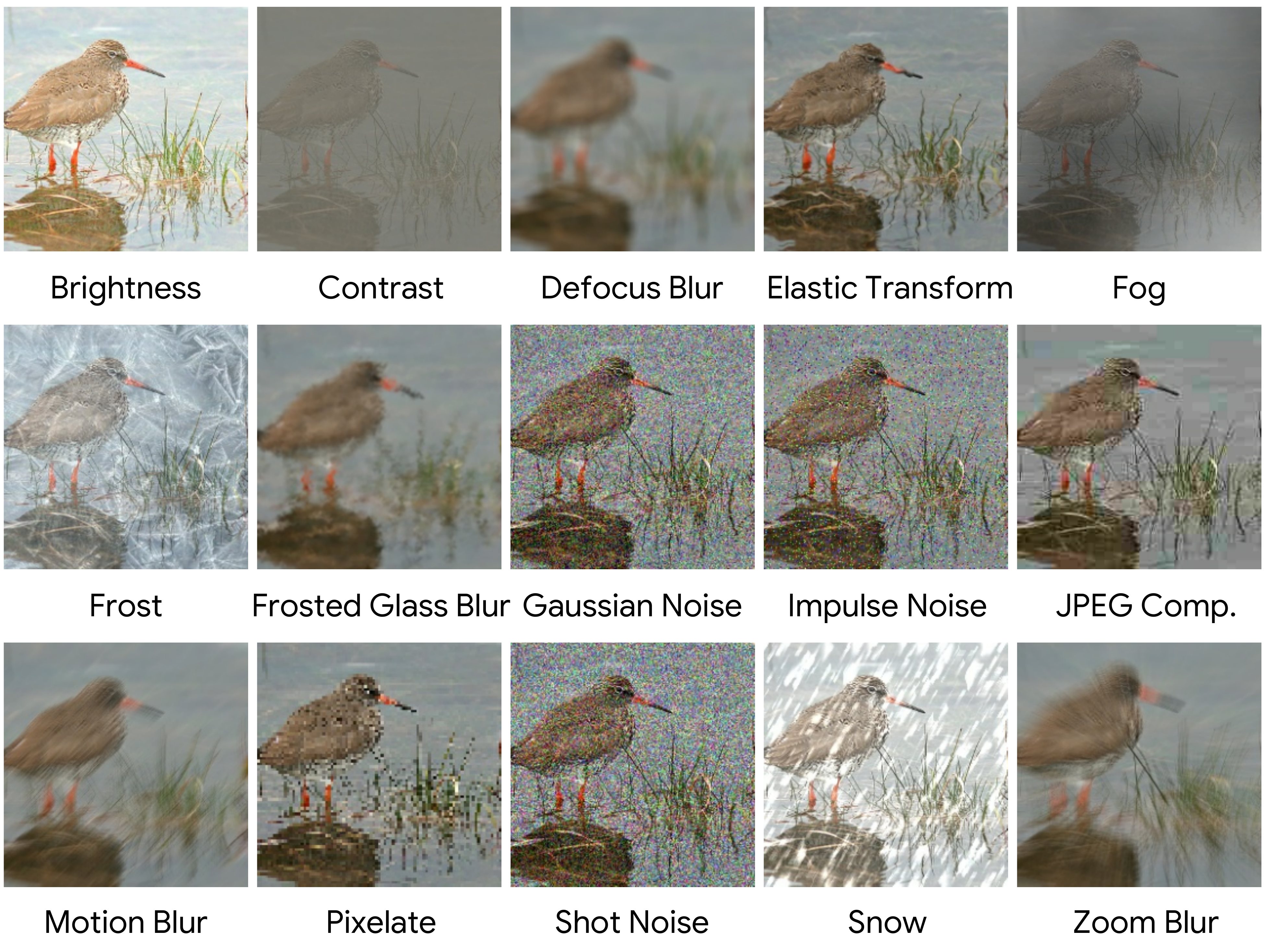}
    \caption{15 Corruption types for the ImageNet-C dataset. All corruptions are shown at severity level 3.}
    \label{fig:imagenet_corruptions}
\end{figure*}

{\flushleft \bf{ImageNet-C:}} The ImageNet-C dataset consists of perturbed and corrupted versions of the ImageNet validation set images. The ImageNet-C data consists of 15 different modes of corruption which are shown in Figure~\ref{fig:imagenet_corruptions}. These corruptions are algorithmically generated from 4 broad categories -- \emph{noise}, \emph{blur}, \emph{weather} and \emph{digital}. Each type of corruption has 5 levels of severity, resulting in 75 distinct corruptions. These corruption types at different severity levels simulate a wide variety of challenging and realistic test time distribution shifts. Each corruption set has $50,000$ test images, which are at $224 \times 224$ resolution. For the results in the paper, we report the mean Classification Accuracy on all 15 corruptions of the highest severity level (level $5$), with results for other severity levels available in Appendix~\ref{app:sev_levels}.

{\flushleft \bf{CIFAR-10-C:}} The CIFAR-10-C dataset consists of perturbed and corrupted versions of the CIFAR-10 test set images. The CIFAR-10-C data consists of 15 different modes of corruption, which are the same as in the ImageNet-C case. Similar to ImageNet-C, each of the corruption types has 5 different levels of severity, resulting in 75 distinct corruptions. Each corruption test set consists of $10,000$ images, which are at $32 \times 32$ resolution.  For the results in the paper, we report the mean Classification Accuracy on all 15 corruptions of the highest severity level (level $5$), as is done in other recent works~\cite{Wang_ICLR_2021, Sun_ICML_2020}. Results for other severity levels are available in Appendix~\ref{app:sev_levels}. 

{\flushleft \bf{ImageNet-A:}} The ImageNet-A dataset~\cite{hendrycks2019nae} consists of naturally occurring adversarial examples from the ImageNet label space that ResNet-50 models fail to classify. The dataset has $7500$ examples.

{\flushleft \bf{ImageNet-R:}} The ImageNet-R dataset~\cite{hendrycks2020many} consists of art, cartoons, graffiti, embroidery, graphics, origami, paintings, patterns, plastic objects, plush objects, sculptures, sketches, tattos, toys, and video game renditions of 200 ImageNet classes. The label space of the data is same as the ImageNet 1000 classes, however, the 200 classes for which renditions are generated are known. The dataset has $30000$ examples.

\section{Qualitative Results}
% add the first result as well

\begin{figure*}[t]
   
     \centering
     \begin{subfigure}[t]{1.0\textwidth}
        \centering
        \includegraphics[scale=0.25]{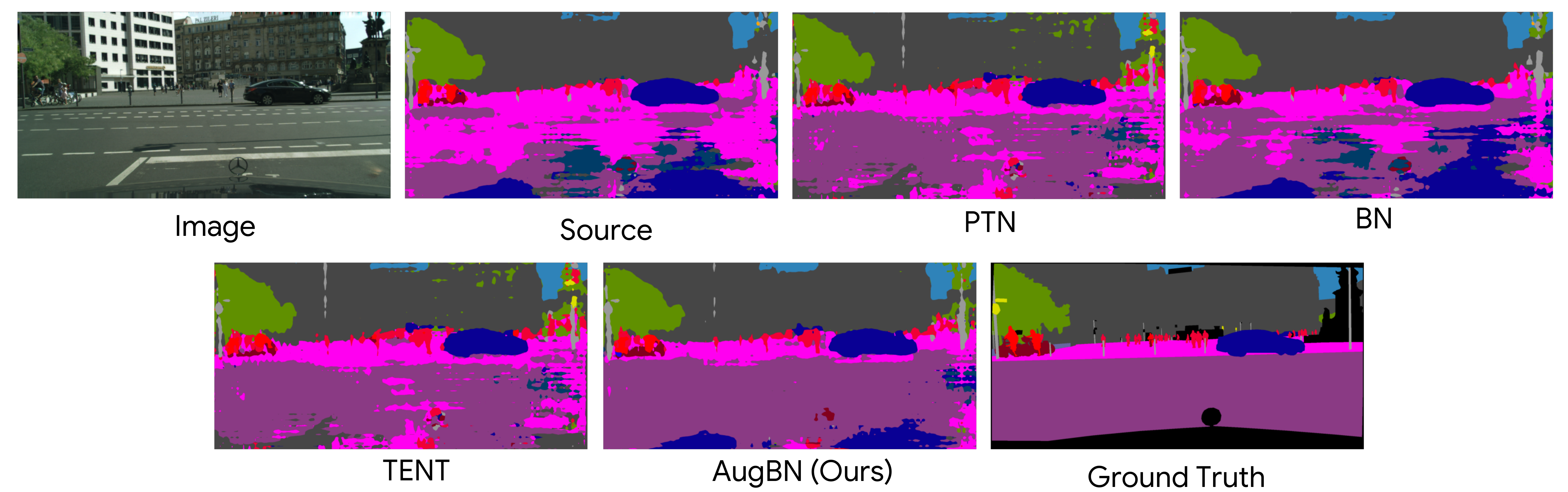}
        \caption{}
        \label{fig:segmentation_qual_ex0}
     \end{subfigure}
     \begin{subfigure}[t]{1.0\textwidth}
         \centering
         \includegraphics[scale=0.25]{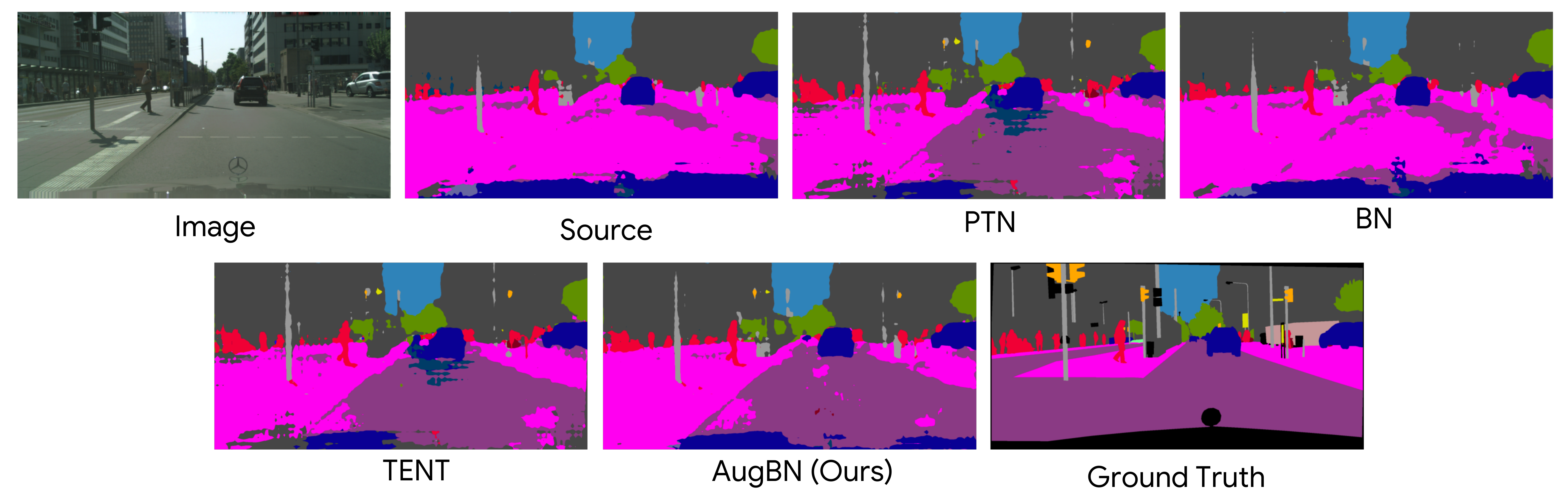}
         \caption{}
         \label{fig:segmentation_qual_ex1}
     \end{subfigure}
     \begin{subfigure}[t]{1.0\textwidth}
         \centering
         \includegraphics[scale=0.25]{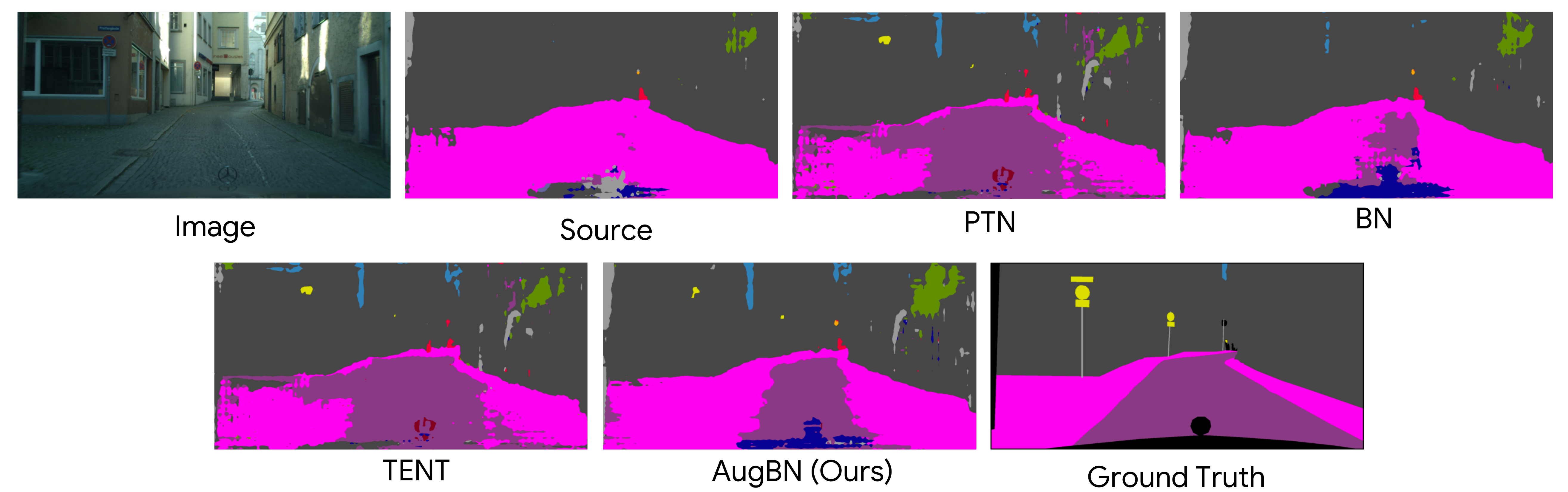}
         \caption{}
         \label{fig:segmentation_qual_ex2}
     \end{subfigure}
    
     \caption{Additional qualitative results for the Semantic Segmentation task.}
    \label{fig:segmentation_qual}
\end{figure*}

Figure \ref{fig:segmentation_qual} presents three examples of the segmentation outputs for all the methods. We observe that the segmentation quality is much better in ``AugBN" than ``Source", as the latter confuses road with sidewalk for large portions of the image in all examples, whereas after calibrating the batch-normalization statistics using AugBN, the predictions are significantly more aligned with the ground-truth. In Figure~\ref{fig:segmentation_qual_ex1} the persons and cars are better defined in AugBN results when compared to other methods.

\begin{figure}[htbp]
    \centering
    \includegraphics[scale=0.3]{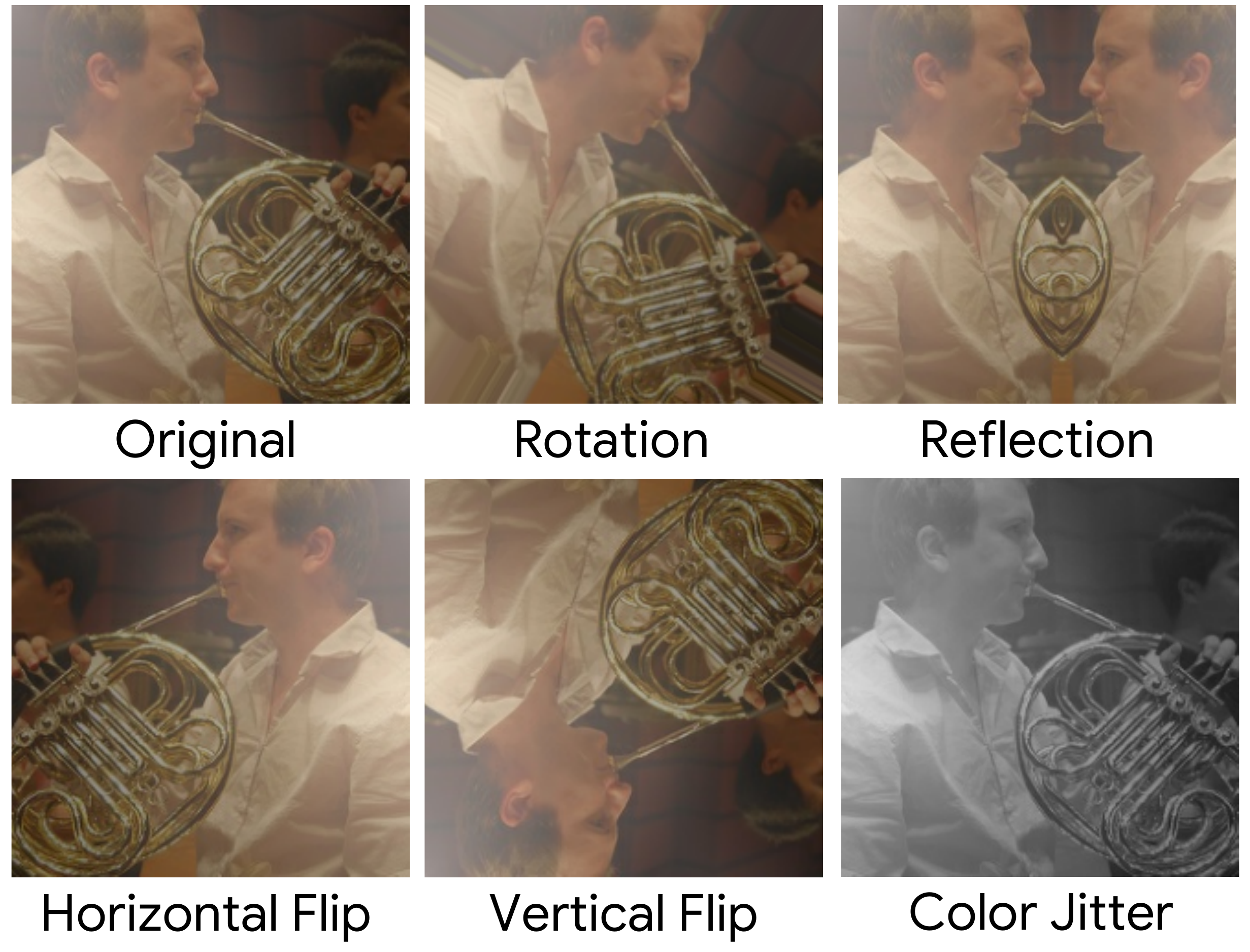}
    \caption{Illustration of augmentations used to create a batch with augmented samples.
    }
    \label{fig:augmentations}
\end{figure}

\begin{table}[!t]
% \begin{wraptable}{l}{5.5cm}
            \small
			\caption{
			Leave-One-Out study of augmentations on CIFAR-10-C.
    	    }
			\label{table:tta_augment_ablation_eccv}
			\centering
			\renewcommand{\arraystretch}{1.2}
	       % \setlength{\tabcolsep}{4pt}
	       % \resizebox{0.8\textwidth}{!}{
			\begin{tabular}{lccccc}
		    \toprule
		  %  \shortstack{Augments $\rightarrow $ \\ Method $\downarrow$}
		    Augments & \shortstack{Color \\ Jitter} & Rotate & Reflect & \shortstack{Horizontal \\ Flip} &  \shortstack{Vertical \\ Flip} \\
		    \midrule
		  %  Select-One & 71.3 & 68.4 & 70.5 & 67.1 & 68.8 &  72.8 & 64.6 \\
		    mCA & 71.8 & 72.7 & 71.5 & 71.9 & 68.8 \\
		  %  \hline
		    \bottomrule
	    \end{tabular}
	    
	   % }
	   % \vspace{-15pt}
	    
\end{table}

\section{Augmentations}
\label{app:augmentations}
We show various augmentations  $\{a_1, \dots, a_m\}$ from which augmented samples are generated for the AugBN algorithm in Figure~\ref{fig:augmentations}. Out of these augmentations, we randomly pick $k (\leq m)$ augmentations and use them as described in Algorithm 2 in the paper.
% {\color{blue} 
For segmentation, we use $k=m=2$ and for classification we use $k=m=5$.
% } 
% {\color{red}
Since our approach samples a set of augmentations randomly from a large set of diverse augmentations, our results do not rely on a specific choice of augmentations. We repeat our experiments with 5 different random seeds to obtain $71.93 \pm 0.04$ mCA with AugBN and $72.94 \pm 0.08$ with AugBN+OPS on CIFAR-10-C.
% To illustrate this, 
We perform a leave-one-out ablation study for the choice of augmentations where we remove each augment from the set of possible augments and test the performance of our AugBN method. We note that the performance of the method remains stable across different augmentations as seen in Table ~\ref{table:tta_augment_ablation_eccv}. The performance with AugBN is better than Source ($61.4$ mCA) and best baseline BN~\cite{Schneider_Neurips_2020} ($65.1$ mCA) for all leave-one-out ablations.
% }

\begin{figure}[htbp]
    \centering
    \includegraphics[scale=0.55]{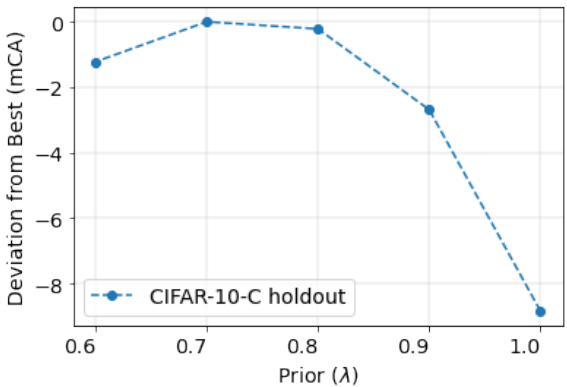}
    \caption{Performance of AugBN (relative to the maximum performance obtained) with varying $\lambda$ on the CIFAR-10-C holdout corruptions. We used $\lambda=0.7$ for the CIFAR-10-C classification results in the paper. 
    %This plot presents the performance of our algorithm on the three datasets. For each dataset, the results are relative to the maximum performance obtained for that dataset. 
    }
    \label{fig:cifar10_momentum}
\end{figure}

\begin{figure}[htbp]
    \centering
    \includegraphics[scale=0.55]{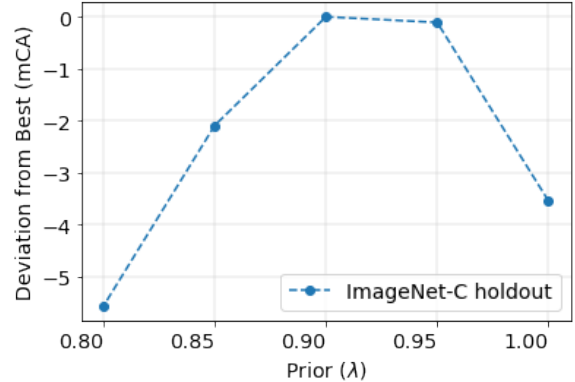}
    \caption{Performance of AugBN (relative to the maximum performance obtained) with varying $\lambda$ on the CIFAR-10-C holdout corruptions. We used $\lambda=0.9$ for the ImageNet-C classification results in the paper.
    }
    \label{fig:imagenet_momentum}
\end{figure}

\begin{figure}[htbp]
    \centering
    \includegraphics[scale=0.55]{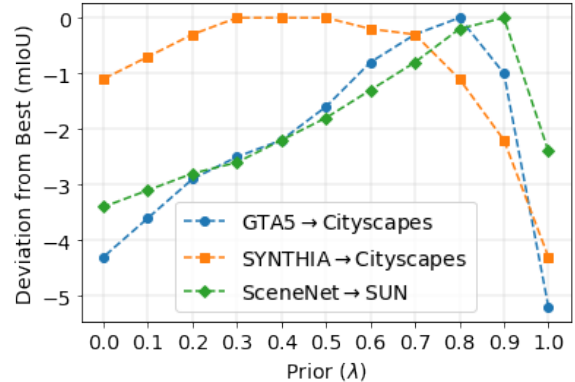}
    \caption{Performance of AugBN (relative to the maximum performance obtained) with varying $\lambda$. We used $\lambda=0.8$ for all the segmentation experiments.
    %This plot presents the performance of our algorithm on the three datasets. For each dataset, the results are relative to the maximum performance obtained for that dataset. 
    }
    \label{fig:segmentation_momentum}
\end{figure}

\begin{figure*}[htbp!]
    \centering
    \includegraphics[trim={1cm, 3cm, 1.2cm, 2cm}, clip, scale=0.7]{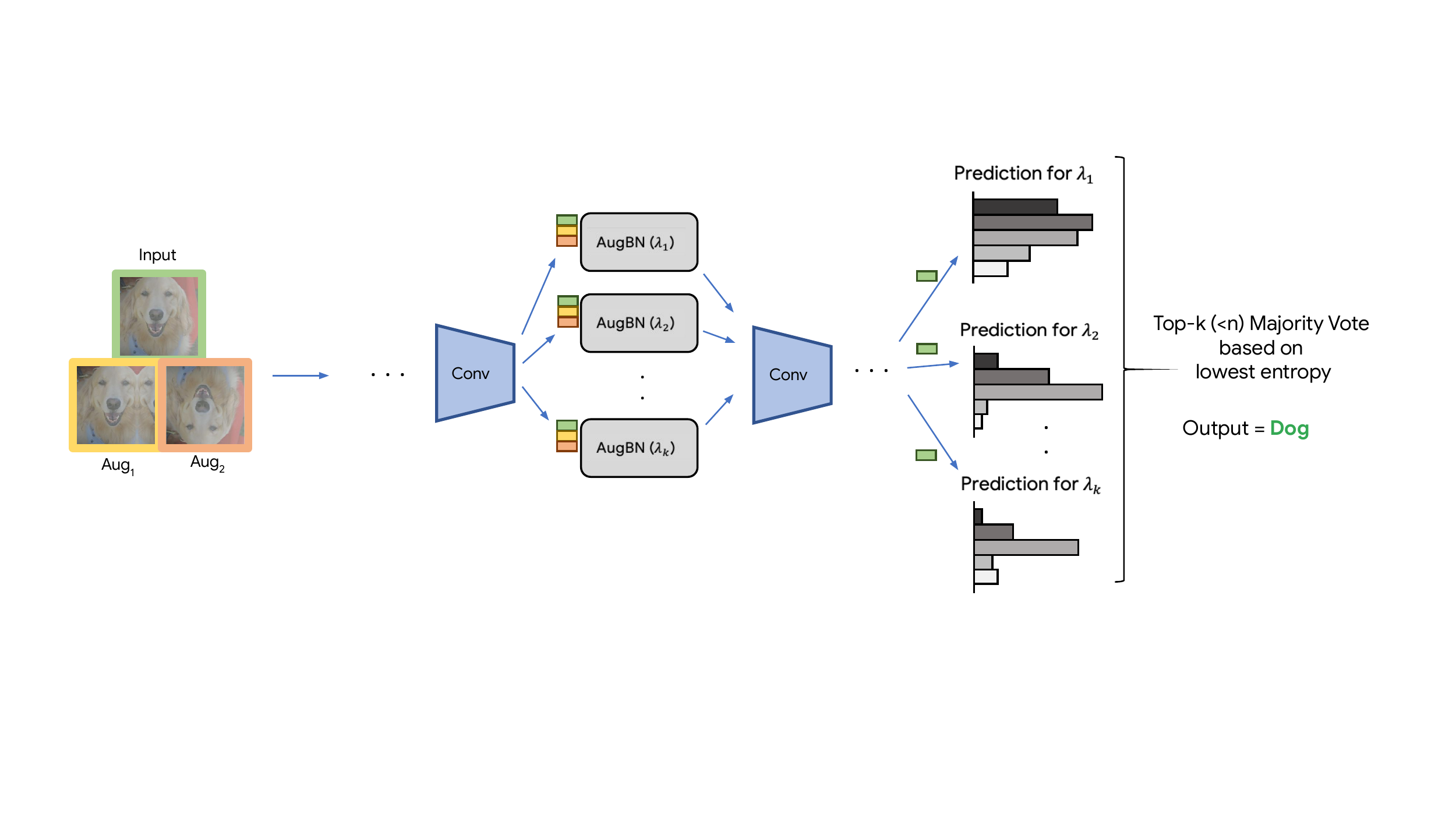}
    \caption{Illustration for forward pass with our AugBN+OPS method. Each AugBN layer maintains a vector of statistics for each prior choice. The model returns the final prediction based on a majority vote on the entropy of the output distribution.}
    \label{fig:augbn_ops}
\end{figure*}

\section{Choice of the Prior $\lambda$}

All test time adaptation methods ~\cite{Sun_ICML_2020, Wang_ICLR_2021, Mummadi_Arxiv_2021, Schneider_Neurips_2020, You_Arix_2021, Zhang_Arxiv_2021, Hu_Arxiv_21} rely on hyper-parameters which are picked beforehand and used during the adaptation process. While we do not expect a holdout or validation set to tune adaptation parameters in the realistic test time adaptation setting, these can be used if they are available. ImageNet-C and CIFAR-10-C datasets have 4 holdout corruption sets - \emph{Spatter}, \emph{speckle noise}, \emph{saturate} and \emph{Gaussian blur}. The method can be evaluated on these holdout sets to get a reasonable indication of which range of $\lambda$ values which can work well in practice. For CIFAR-10-C, the best mCA of $74.9$ is obtained at $\lambda=0.7$, and remains stable for $\sim 20\%$ of the prior range, as shown in Figure~\ref{fig:cifar10_momentum}. For ImageNet-C, the best mCA of $31.8$ is at $\lambda=0.9$, and the deviation from the best result at different priors ($\lambda$) is shown in Figure~\ref{fig:imagenet_momentum}. We use these prior values for the results in the paper. However, in the case of segmentation, there are no holdout sets, so we use a fixed prior $\lambda = 0.8$ for all datasets, even when better results were available at other $\lambda$ values. For example, in Figure~\ref{fig:segmentation_momentum}, $\lambda = 0.9$ produces a better result for the SceneNet $\rightarrow$ SUN and $\lambda=0.3$ perform the best for SYNTHIA $\rightarrow$ Cityscapes, but we still report results with $\lambda = 0.8$ in the paper, being consistent with the spirit of test time adaptation. Further, these results demonstrate that the results are not very sensitive with respect to the exact choice of $\lambda$.

\begin{figure}[!htbp]
    \centering
    \hspace{-2mm}
    \includegraphics[clip, trim={2.5cm 0 6cm 3.2cm}, scale=0.5]{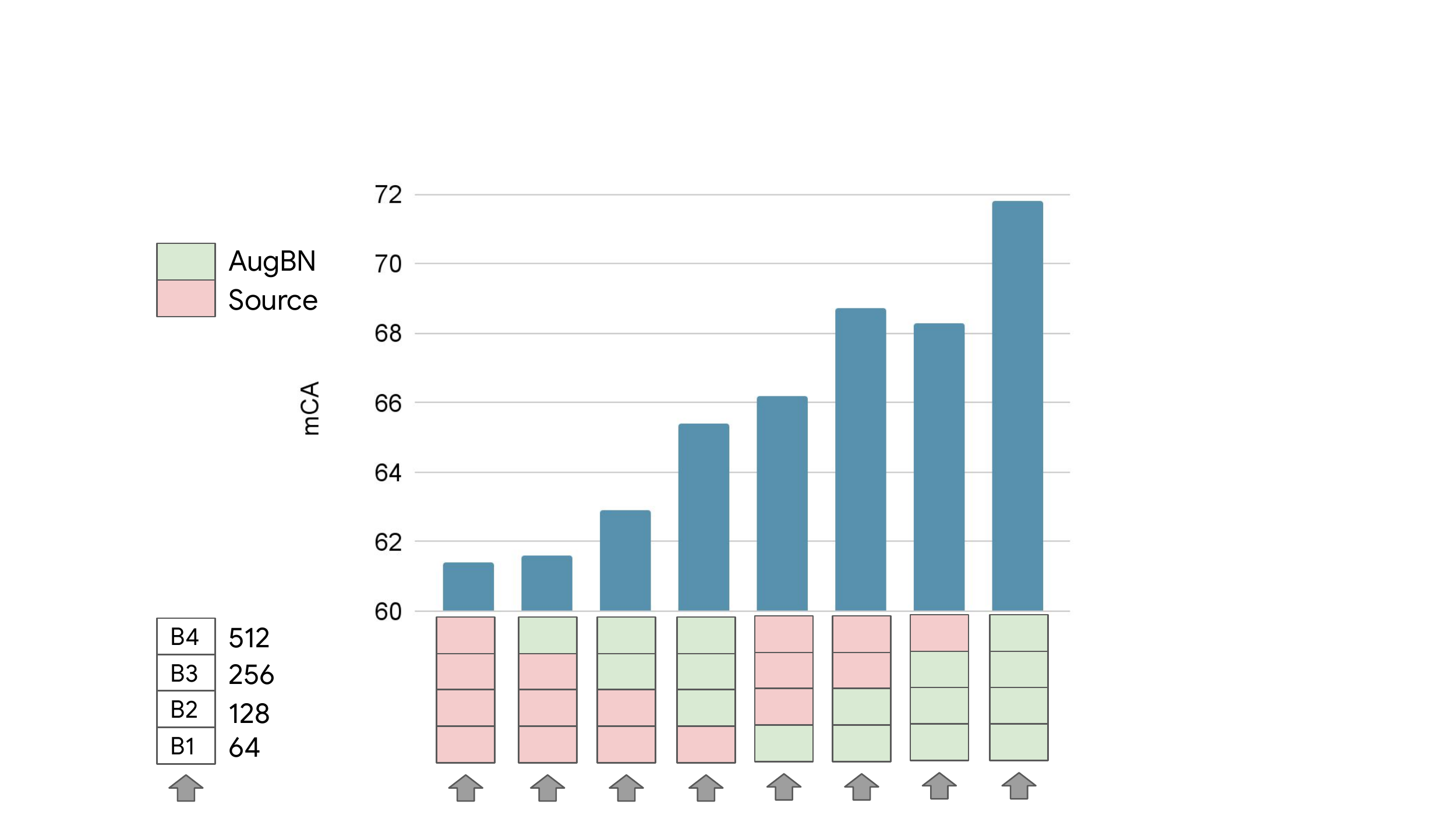}
    \vspace{-1em}
    \caption{Results for layer-wise sensitivity analysis and the role of AugBN in performance improvement.} 
    \label{fig:sensitivity}
    \vspace{-16pt}
\end{figure}

\begin{figure*}[!htbp]
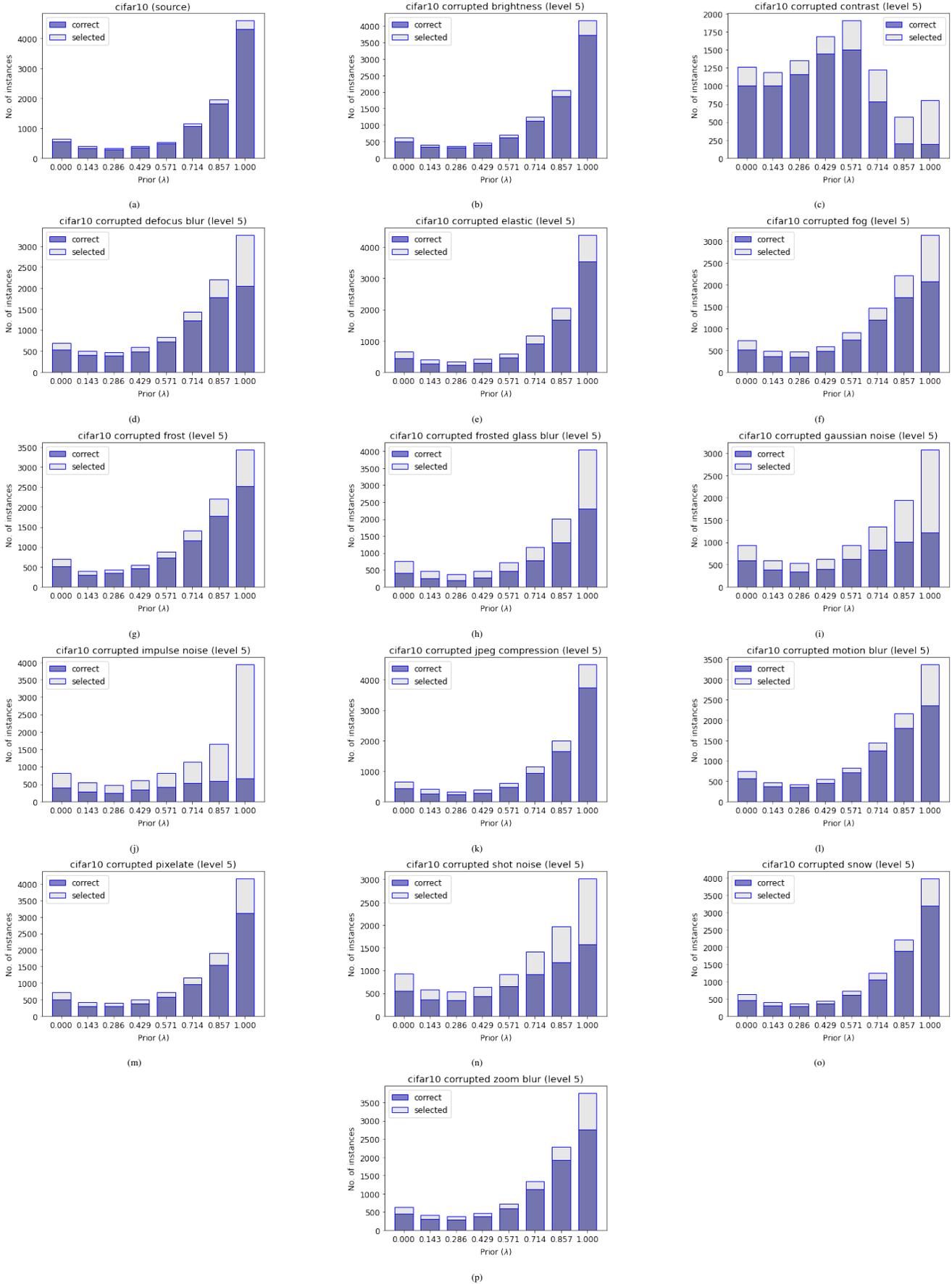

\centering
\captionsetup{font=scriptsize}
\begin{subfigure}[h]{0.28\textwidth}
         \centering
         \includegraphics[scale=1, width=\textwidth]{figures/lambda_selection_correct/cifar10_source_.pdf}
         \caption{}
\end{subfigure}
\hfill
\begin{subfigure}[h]{0.28\textwidth}
         \centering
         \includegraphics[scale=1, width=\textwidth]{figures/lambda_selection_correct/cifar10_corrupted_brightness_level_5_.pdf}
         \caption{}
\end{subfigure}
\hfill
\begin{subfigure}[h]{0.28\textwidth}
         \centering
         \includegraphics[scale=1, width=\textwidth]{figures/lambda_selection_correct/cifar10_corrupted_contrast_level_5_.pdf}
         \caption{}
\end{subfigure}

\begin{subfigure}[h]{0.28\textwidth}
         \centering
         \includegraphics[scale=1, width=\textwidth]{figures/lambda_selection_correct/cifar10_corrupted_defocus_blur_level_5_.pdf}
         \caption{}
\end{subfigure}
\hfill
\begin{subfigure}[h]{0.28\textwidth}
         \centering
         \includegraphics[scale=1, width=\textwidth]{figures/lambda_selection_correct/cifar10_corrupted_elastic_level_5_.pdf}
         \caption{}
\end{subfigure}
\hfill
\begin{subfigure}[h]{0.28\textwidth}
         \centering
         \includegraphics[scale=1, width=\textwidth]{figures/lambda_selection_correct/cifar10_corrupted_fog_level_5_.pdf}
         \caption{}
\end{subfigure}

\begin{subfigure}[h]{0.28\textwidth}
         \centering
         \includegraphics[scale=1, width=\textwidth]{figures/lambda_selection_correct/cifar10_corrupted_frost_level_5_.pdf}
         \caption{}
\end{subfigure}
\hfill
\begin{subfigure}[h]{0.28\textwidth}
         \centering
         \includegraphics[scale=1, width=\textwidth]{figures/lambda_selection_correct/cifar10_corrupted_frosted_glass_blur_level_5_.pdf}
         \caption{}
\end{subfigure}
\hfill
\begin{subfigure}[h]{0.28\textwidth}
         \centering
         \includegraphics[scale=1, width=\textwidth]{figures/lambda_selection_correct/cifar10_corrupted_gaussian_noise_level_5_.pdf}
         \caption{}
\end{subfigure}

\begin{subfigure}[h]{0.28\textwidth}
         \centering
         \includegraphics[scale=1, width=\textwidth]{figures/lambda_selection_correct/cifar10_corrupted_impulse_noise_level_5_.pdf}
         \caption{}
\end{subfigure}
\hfill
\begin{subfigure}[h]{0.28\textwidth}
         \centering
         \includegraphics[scale=1, width=\textwidth]{figures/lambda_selection_correct/cifar10_corrupted_jpeg_compression_level_5_.pdf}
         \caption{}
\end{subfigure}
\hfill
\begin{subfigure}[h]{0.28\textwidth}
         \centering
         \includegraphics[scale=1, width=\textwidth]{figures/lambda_selection_correct/cifar10_corrupted_motion_blur_level_5_.pdf}
         \caption{}
\end{subfigure}

\begin{subfigure}[h]{0.28\textwidth}
         \centering
         \includegraphics[scale=1, width=\textwidth]{figures/lambda_selection_correct/cifar10_corrupted_pixelate_level_5_.pdf}
         \caption{}
\end{subfigure}
\hfill
\begin{subfigure}[h]{0.28\textwidth}
         \centering
         \includegraphics[scale=1, width=\textwidth]{figures/lambda_selection_correct/cifar10_corrupted_shot_noise_level_5_.pdf}
         \caption{}
\end{subfigure}
\hfill
\begin{subfigure}[h]{0.28\textwidth}
         \centering
         \includegraphics[scale=1, width=\textwidth]{figures/lambda_selection_correct/cifar10_corrupted_snow_level_5_.pdf}
         \caption{}
\end{subfigure}

\begin{subfigure}[h]{0.28\textwidth}
         \centering
         \includegraphics[scale=1, width=\textwidth]{figures/lambda_selection_correct/cifar10_corrupted_zoom_blur_level_5_.pdf}
         \caption{}
\end{subfigure}
\caption{Distribution of choice of Prior $(\lambda)$ (on CIFAR-10-C datasets) based on the minimum entropy criterion. The fraction of images which were correctly classified using the chosen prior are shown by the highlighted regions within each bar. $(\lambda)=1$ indicates using the source statistics while $(\lambda)=0$ indicates using the statistics inferred from the given test sample.}
\label{fig:cifar10_lambda_choice}
\end{figure*}
% \clearpage
\begin{figure*}[!htbp]
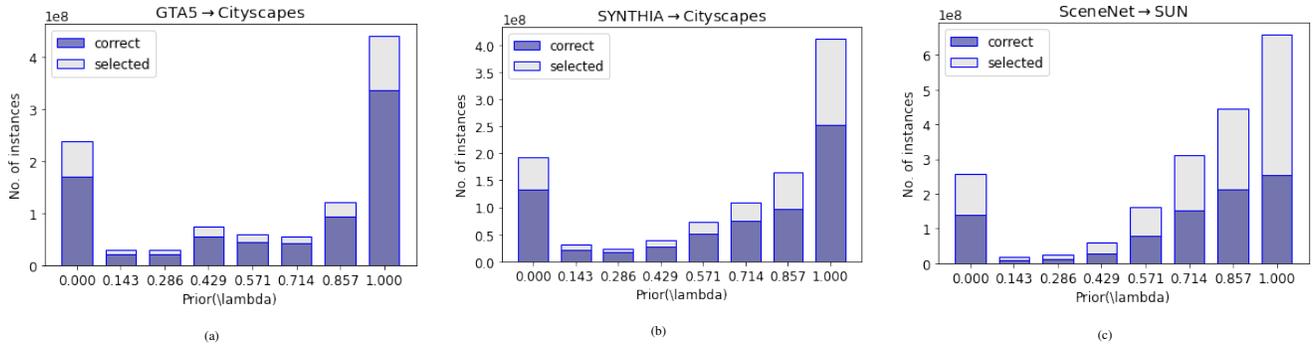

\centering
\captionsetup{font=scriptsize}
\begin{subfigure}[h]{0.32\textwidth}
         \centering
         \includegraphics[scale=1, width=\textwidth]{figures/lambda_selection_correct/gta5_to_cityscapes.pdf}
         \caption{}
\end{subfigure}
\hfill
\begin{subfigure}[h]{0.32\textwidth}
         \centering
         \includegraphics[scale=1, width=\textwidth]{figures/lambda_selection_correct/synthia_to_cityscapes.pdf}
         \caption{}
\end{subfigure}
\hfill
\begin{subfigure}[h]{0.32\textwidth}
         \centering
         \includegraphics[scale=1, width=\textwidth]{figures/lambda_selection_correct/scenenet_to_sun.pdf}
         \caption{}
\end{subfigure}

\caption{Choice of Prior $(\lambda)$ (on segmentation datasets) based on the minimum entropy criterion. The fraction of pixels which were correctly classified using the chosen prior are shown by the highlighted regions within each bar. $(\lambda)=1$ indicates using the source statistics while $(\lambda)=0$ indicates using the statistics inferred from the given test sample.}
\label{fig:segmentation_lambda_choice}
\end{figure*}

\subsection{Doing away with a pre-determined $\lambda$}
% {\color{red}
With AugBN+OPS, we do away with the limitation of choosing priors in an ad-hoc manner. AugBN+OPS automatically selects the prior value from a choice of  $n_p$ different priors as discussed in Section 3 of the main paper. For both classification and segmentation, we choose from a set of $n_p=8$ priors and perform majority voting for top-k $(k=3)$ priors based on the minimum entropy criterion. The forward pass with AugBN+OPS is illustrated in Figure~\ref{fig:augbn_ops}. To demonstrate the effectiveness of picking priors based on this criterion, we demonstrate the distribution of test images with respect to the priors selected based on minimum entropy and the fraction of the images which were classified correctly in Figure ~\ref{fig:cifar10_lambda_choice} on the CIFAR-10-C dataset. We further show the pixel-wise choice of prior on the segmentation datasets in Figure ~\ref{fig:segmentation_lambda_choice}. In Figure~\ref{fig:cifar10_lambda_choice}, for the source test set, prior is chosen as $\lambda=1.0$ for most samples. For harder corruptions, like Contrast, the distribution of chosen priors weighs more towards $\lambda=0.5$, such that more importance is given to the current test estimate $(\mu_t)$. The chosen priors $(\lambda)$ show similar fractions of correct images/pixels, thus denoting that chosen prior values are contributing to the overall performance for each test set.  Note that these plots show selection of prior based on the lowest entropy value. The final results perform a majority voting over the top-k $(k=3)$ priors based on the lowest entropy. 
% To the best of our knowledge, we are the first to propose an hyperparameter-free test time adaptation procedure.
% }
% \AK{Add discussion that AugBN+OPS does not need selection of prior. Add plots for CIFAR-10 and choice of augments.}

\section{Layer-wise analysis of AugBN}

To study how AugBN directly impacts the model performance and which layers are the most affected by AugBN, we design the following experiment: For results on CIFAR-10-C, we use ResNet-26, which has 4 block-groups, grouped by number of features (64, 128, 256, 512), each consisting of 3 Residual blocks. 
%Each Residual block is a combination of 2 Conv and 2 BN layers with a shortcut branch.
%To evaluate the impact of AugBN layers, 
We selectively replace Source BN layers with AugBN layers for each Residual Block and observe the performance (Figure~\ref{fig:sensitivity}). 
%The results for all configurations are shown in Figure~\ref{fig:sensitivity}. 
First column represents all blocks using source BN layers, and the last column represents our result using AugBN in all the layers. In columns 2-4, we set the initial blocks to gradually use Source BN and in columns 5-7 we set the initial blocks to use AugBN. We observe that the first layer is the most important for capturing domain specific information. With only the initial block using AugBN (column 5), we obtain higher mCA as compared to using Source statistics in the first block followed by AugBN on all subsequent blocks. This also demonstrates how good single image statistic estimates can directly boost performance.

\begin{table}[htbp!]
            % \vspace{-0.3cm}
			\caption{ImageNet-C results on additional architectures.
    	    }
			\label{table:add_arch_imagenetc}
			\centering
			\renewcommand{\arraystretch}{1.2}
	        \setlength{\tabcolsep}{4pt}
			\begin{tabular}{lcc}
		    \toprule
		    Methods & DenseNet121 & InceptionV3 \\
		    \midrule
		    % \multicolumn{4}{c}{0 Iteration} \\
		    Source & 22.7 & 30.4 \\
		    PTN & 0.1 & 0.2 \\
		    BN & 25.6 & 35.8\\
		    \midrule
		    AugBN & 25.7 & 36.0 \\
		    AugBN+OPS & \textbf{25.8} & \textbf{36.2}\\
		    \bottomrule
	    \end{tabular}
\end{table}

\begin{table}[htbp!]
            % \vspace{-0.3cm}
			\caption{ImageNet-A results on additional architectures.
    	    }
			\label{table:add_arch_imageneta}
			\centering
			\renewcommand{\arraystretch}{1.2}
	        \setlength{\tabcolsep}{4pt}
			\begin{tabular}{lcc}
		    \toprule
		    Methods & DenseNet121 & InceptionV3 \\
		    \midrule
		    % \multicolumn{4}{c}{0 Iteration} \\
		    Source & 0.7 & 3.2 \\
		    PTN & 0 & 0.1 \\
		    BN & 1.3 & 3.5\\
		    \midrule
		    AugBN & 1.2 & \textbf{3.8} \\
		    AugBN+OPS & \textbf{1.4} & 3.7\\
		    \bottomrule
	    \end{tabular}
\end{table}

\begin{table}[htbp!]
            % \vspace{-0.3cm}
			\caption{ImageNet-R results on additional architectures.
    	    }
			\label{table:add_arch_imagenetr}
			\centering
			\renewcommand{\arraystretch}{1.2}
	        \setlength{\tabcolsep}{4pt}
			\begin{tabular}{lcc}
		    \toprule
		    Methods & DenseNet121 & InceptionV3 \\
		    \midrule
		    % \multicolumn{4}{c}{0 Iteration} \\
		    Source & 37.5 & 39.1 \\
		    PTN & 0.6 & 1 \\
		    BN & 40.2 & 42.8\\
		    \midrule
		    AugBN & 40.9 & 43.9 \\
		    AugBN+OPS & \textbf{41.4} & \textbf{46.0}\\
		    \bottomrule
	    \end{tabular}
\end{table}

% \AK{Add rebuttal experiments}
\section{Results on additional architectures}
\label{app:architectures}
To showcase the universality of our approach we tested our method on pre-trained DenseNet-121 and InceptionV3 architectures trained on the ImageNet dataset. We show results for ImageNet-C, ImageNet-A and ImageNet-R in Tables ~\ref{table:add_arch_imagenetc}, ~\ref{table:add_arch_imageneta} and ~\ref{table:add_arch_imagenetr} respectively. Our approach results in significant performance gains across various datasets and model architectures. 

\begin{table}[!t]
            \small
			\caption{Results on CIFAR10-C and ImageNet-C datasets. We report the mean Classification Accuracy (mCA) over all 15 corruptions of the highest severity (level 5). $^*$MixNorm~\cite{Hu_Arxiv_21} and MEMO~\cite{Zhang_Arxiv_2021} are unpublished works on Arxiv and their performance is reported as is. MixNorm~\cite{Hu_Arxiv_21} does not report their source model performance.
    	    }
			\label{table:sota_concurrent}
			\centering
			\renewcommand{\arraystretch}{1.2}
	        \setlength{\tabcolsep}{4pt}
			\begin{tabular}{lcc}
		    \toprule
		    Methods & CIFAR-10-C & ImageNet-C\\
		    \midrule
		    % \multicolumn{4}{c}{0 Iteration} \\
		    Source (MEMO) & 67.3 & 18.0 \\
		    MEMO$^*$ & 70.3 & 24.2 \\
		    MixNorm$^*$ & 68.0 & 21.1 \\
		    \midrule
		    Source & 61.4 & 20.5 \\
		    AugBN & 71.8 & 25.0 \\
		    AugBN+OPS & \textbf{73.1} & \textbf{25.5}\\
		    \bottomrule
	    \end{tabular}
\end{table}

\section{Results at all severity levels}
\label{app:sev_levels}
\begin{figure*}[htbp]
    % \hspace{-3mm}
    \centering
    \includegraphics[scale=0.5]{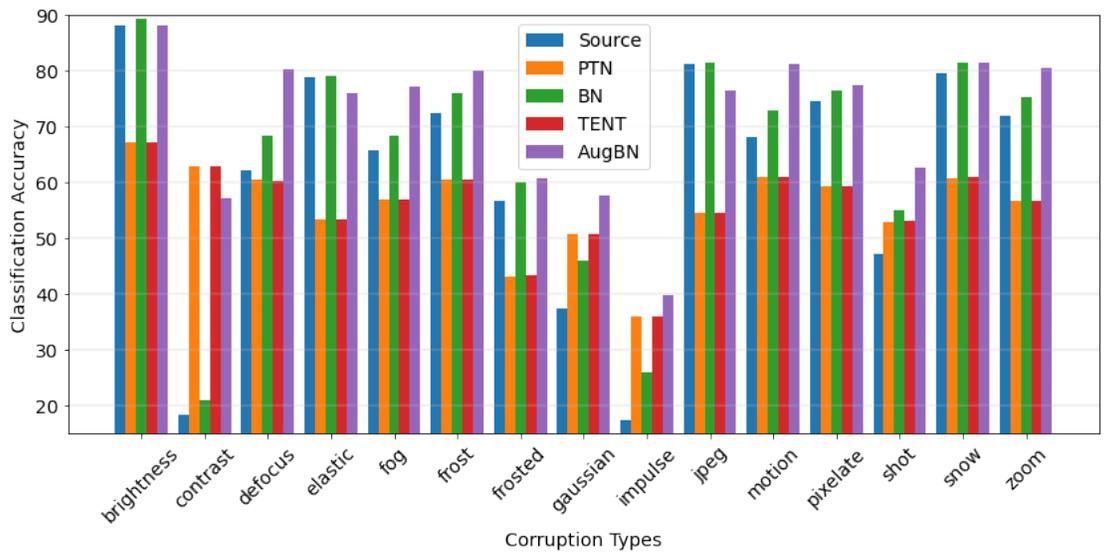}
    \caption{Performance comparison on CIFAR10-C corruptions.}
    \label{fig:cifar10_corruptions}
\end{figure*}

{\flushleft \bf{CIFAR-10-C:}} We show classification accuracy for all 5 severity levels, for each of the 15 corruption types for all the methods, in Table \ref{table:cifar10_sev1}-\ref{table:cifar10_sev5}. At lower corruption levels, the performance of the source model itself is very good, as the source distribution is close to the the test distribution. We do not tune the augmentations for different corruption levels.
Emphasis given to the augmented samples can be controlled by the prior ($\lambda$). We use $\lambda = 0.9$ for severity levels 1-3, and $\lambda = 0.7$ for higher severity levels. AugBN provides significant gains in performance over the source model in all cases, especially at high severity level corruptions (Table~\ref{table:cifar10_sev5}). 
% {\color{red} 
Corruption wise performance on severity level 5 is shown in Figure~\ref{fig:cifar10_corruptions}.
% }

{\flushleft \bf{ImageNet-C:}} We show classification accuracy for all 5 severity levels, for each of the 15 corruption types for all the methods, in Table \ref{table:imagenet_sev1}-\ref{table:imagenet_sev5}. At lower corruption levels, the prior ($\lambda$) needs to have more emphasis on the source statistics. We use $\lambda = 0.95$ for severity levels 1 and 2, and use $\lambda = 0.9$ for the remaining cases. AugBN provides significant gains in performance over the source model, especially at high severity level corruptions (Table~\ref{table:imagenet_sev5}).

\clearpage
\begin{table*} [htbp]
	\caption{Results for CIFAR-10-C dataset, with severity level 1 corruptions.
	}
% 	\vspace{1mm}
	\label{table:cifar10_sev1}
% 	\scriptsize
	\centering
	\renewcommand{\arraystretch}{1.7}
	\resizebox{\textwidth}{!}{
	\begin{tabular}{lccccccccccccccccc}
		\toprule
		
		%& & \multicolumn{20}{c}{GTA5 $\rightarrow$ Cityscapes} \\
		%\midrule
		
Method & brightness & contrast & defocus blur & elastic & fog & frost & frosted glass blur & gaussian noise & impulse noise & jpeg compression & motion blur & pixelate & shot noise & snow & zoom blur & mCA  \\ 
\midrule
Source & 93.1 & 92.1 & 93.2 & 89.3 & 92.7 & 91.1 & 65.5 & 87.4 & 78.5 & 89.9 & 90.0 & 92.2 & 90.3 & 88.6 & 87.5 & 88.1  \\ 
PTN~\cite{Nado_Arxiv_2020, Schneider_Neurips_2020} & 69.6 & 69.6 & 69.5 & 65.0 & 69.7 & 67.9 & 50.6 & 65.4 & 57.8 & 63.8 & 67.5 & 68.1 & 67.6 & 63.9 & 65.8 & 65.5  \\ 
BN~\cite{Schneider_Neurips_2020} & 93.4 & 92.5 & 93.2 & 89.7 & 92.9 & 91.6 & 69.2 & 88.6 & 80.6 & 90.1 & 90.7 & 92.4 & 90.8 & 88.9 & 88.3 & 88.9  \\ 
TENT~\cite{Wang_ICLR_2021} & 69.9 & 69.8 & 69.5 & 65.2 & 69.7 & 68.3 & 50.5 & 65.6 & 58.1 & 63.9 & 67.5 & 68.3 & 67.8 & 64.0 & 65.8 & 65.6  \\ 
% AugBN & 93.2 & 92.7 & 93.0 & 89.6 & 92.7 & 91.6 & 71.0 & 88.9 & 80.8 & 90.0 & 90.9 & 92.2 & 91.0 & 88.7 & 88.7 &\textbf{ 89.0}  \\ 
AugBN & 93.1 & 92.6 & 93.1 & 89.8 & 92.8 & 91.4 & 69.5 & 88.4 & 80.6 & 89.9 & 91.0 & 92.1 & 90.6 & 88.8 & 89.2 & \textbf{88.9}  \\ 

	\bottomrule
	\end{tabular}
	}
\end{table*}

\begin{table*} [htbp]
	\caption{Results for CIFAR-10-C dataset, with severity level 2 corruptions.
	}
% 	\vspace{1mm}
	\label{table:cifar10_sev2}
% 	\scriptsize
	\centering
	\renewcommand{\arraystretch}{1.7}
	\resizebox{\textwidth}{!}{
	\begin{tabular}{llccccccccccccccccc}
		\toprule
		
Method & brightness & contrast & defocus blur & elastic & fog & frost & frosted glass blur & gaussian noise & impulse noise & jpeg compression & motion blur & pixelate & shot noise & snow & zoom blur & mCA  \\ 
\midrule
Source & 92.6 & 86.3 & 91.7 & 89.3 & 90.9 & 87.5 & 66.8 & 74.0 & 65.0 & 86.2 & 84.4 & 91.1 & 86.0 & 82.5 & 86.0 & 84.0  \\ 
PTN~\cite{Nado_Arxiv_2020, Schneider_Neurips_2020} & 69.4 & 69.7 & 68.9 & 64.4 & 68.4 & 66.4 & 51.5 & 60.8 & 51.8 & 60.0 & 65.9 & 66.6 & 64.8 & 60.3 & 64.7 & 63.6  \\ 
BN~\cite{Schneider_Neurips_2020} & 92.9 & 88.1 & 91.9 & 89.5 & 91.4 & 88.4 & 70.2 & 77.8 & 68.7 & 86.6 & 86.2 & 91.1 & 87.4 & 83.8 & 87.3 & 85.4  \\ 
TENT~\cite{Wang_ICLR_2021} & 69.7 & 69.6 & 69.1 & 64.5 & 68.5 & 66.6 & 51.5 & 61.1 & 51.6 & 60.0 & 65.8 & 66.8 & 64.9 & 60.5 & 64.8 & 63.7  \\ 
% AugBN & 92.2 & 90.0 & 91.1 & 88.8 & 91.0 & 88.7 & 72.5 & 82.3 & 70.1 & 85.2 & 87.3 & 90.1 & 88.2 & 84.3 & 87.5 & \textbf{85.9}  \\ 		
AugBN & 92.8 & 89.4 & 92.0 & 89.9 & 91.6 & 88.7 & 70.4 & 78.4 & 69.2 & 86.7 & 87.5 & 91.1 & 87.1 & 84.1 & 88.5 &\textbf{ 85.8  }\\

	\bottomrule
	\end{tabular}
	}
\end{table*}

\begin{table*} [!htbp]
	\caption{Results for CIFAR-10-C dataset, with severity level 3 corruptions.
	}
% 	\vspace{1mm}
	\label{table:cifar10_sev3}
% 	\scriptsize
	\centering
	\renewcommand{\arraystretch}{1.7}
	\resizebox{\textwidth}{!}{
	\begin{tabular}{lccccccccccccccccc}
		\toprule
Method & brightness & contrast & defocus blur & elastic & fog & frost & frosted glass blur & gaussian noise & impulse noise & jpeg compression & motion blur & pixelate & shot noise & snow & zoom blur & mCA  \\ 
 \midrule 
Source & 91.9 & 78.4 & 88.3 & 86.8 & 87.9 & 81.3 & 72.4 & 55.6 & 54.0 & 85.3 & 76.7 & 90.2 & 69.6 & 83.1 & 82.4 & 78.9  \\ 
PTN~\cite{Nado_Arxiv_2020, Schneider_Neurips_2020} & 69.1 & 69.0 & 67.9 & 63.0 & 67.1 & 63.6 & 52.5 & 56.3 & 47.0 & 58.5 & 63.4 & 65.7 & 59.5 & 60.3 & 62.5 & 61.7  \\ 
BN~\cite{Schneider_Neurips_2020} & 92.2 & 82.0 & 89.3 & 87.3 & 88.8 & 83.4 & 74.1 & 62.5 & 59.8 & 85.7 & 80.3 & 90.5 & 73.7 & 84.1 & 84.3 & 81.2  \\ 
TENT~\cite{Wang_ICLR_2021} & 69.2 & 69.1 & 67.9 & 63.2 & 67.1 & 63.7 & 52.6 & 56.3 & 47.0 & 58.7 & 63.4 & 65.8 & 59.6 & 60.5 & 62.5 & 61.8  \\ 
% AugBN & 92.3 & 83.9 & 89.4 & 87.4 & 89.2 & 84.7 & 75.3 & 67.1 & 61.4 & 85.3 & 81.5 & 90.4 & 76.1 & 84.2 & 84.7 &\textbf{ 82.2 } \\ 
AugBN & 92.3 & 85.0 & 90.2 & 88.0 & 89.5 & 84.1 & 74.2 & 63.8 & 61.0 & 85.5 & 82.7 & 90.3 & 74.1 & 84.2 & 85.8 & \textbf{82.0 } \\ 

	\bottomrule
	\end{tabular}
	}
\end{table*}

\begin{table*} [!htbp]
	\caption{Results for CIFAR-10-C dataset, with severity level 4 corruptions.
	}
% 	\vspace{1mm}
	\label{table:cifar10_sev4}
% 	\scriptsize
	\centering
	\renewcommand{\arraystretch}{1.7}
	\resizebox{\textwidth}{!}{
	\begin{tabular}{lccccccccccccccccc}
		\toprule
Method & brightness & contrast & defocus blur & elastic & fog & frost & frosted glass blur & gaussian noise & impulse noise & jpeg compression & motion blur & pixelate & shot noise & snow & zoom blur & mCA  \\ 
 \midrule 
Source & 91.1 & 61.5 & 81.6 & 82.3 & 82.7 & 80.6 & 51.6 & 45.5 & 30.4 & 83.4 & 76.3 & 85.6 & 61.1 & 81.7 & 78.8 & 71.6  \\ 
PTN~\cite{Nado_Arxiv_2020, Schneider_Neurips_2020} & 68.7 & 68.2 & 65.3 & 57.8 & 64.0 & 62.9 & 43.7 & 53.2 & 39.8 & 56.5 & 63.2 & 64.2 & 56.8 & 59.4 & 60.4 & 58.9  \\ 
BN~\cite{Schneider_Neurips_2020} & 91.4 & 68.1 & 83.8 & 83.3 & 84.5 & 83.3 & 56.2 & 54.0 & 41.6 & 83.8 & 79.8 & 86.5 & 66.6 & 82.5 & 81.3 & 75.1  \\ 
TENT~\cite{Wang_ICLR_2021} & 69.0 & 68.0 & 65.5 & 58.0 & 64.1 & 63.2 & 43.6 & 53.5 & 39.9 & 56.6 & 63.4 & 64.4 & 57.0 & 59.5 & 60.5 & 59.1  \\ 
% AugBN & 89.6 & 82.8 & 84.9 & 79.9 & 84.8 & 84.7 & 60.2 & 68.2 & 53.0 & 79.3 & 82.9 & 85.2 & 75.1 & 79.9 & 82.1 & \textbf{78.2}  \\ 
AugBN & 90.9 & 81.6 & 87.4 & 83.5 & 87.1 & 84.7 & 59.0 & 60.7 & 49.1 & 81.8 & 84.6 & 86.0 & 69.9 & 82.0 & 85.0 & \textbf{78.2}  \\

	\bottomrule
	\end{tabular}
	}
\end{table*}

\begin{table*} [!htbp]
	\caption{Results for CIFAR-10-C dataset, with severity level 5 corruptions.
	}
% 	\vspace{1mm}
	\label{table:cifar10_sev5}
% 	\scriptsize
	\centering
	\renewcommand{\arraystretch}{1.7}
	\resizebox{\textwidth}{!}{
	\begin{tabular}{lccccccccccccccccc}
		\toprule
Method & brightness & contrast & defocus blur & elastic & fog & frost & frosted glass blur & gaussian noise & impulse noise & jpeg compression & motion blur & pixelate & shot noise & snow & zoom blur & mCA  \\ 
 \midrule 
Source & 88.3 & 18.5 & 62.2 & 78.8 & 65.7 & 72.5 & 56.8 & 37.4 & 17.5 & 81.3 & 68.2 & 74.6 & 47.3 & 79.5 & 72.0 & 61.4  \\ 
PTN~\cite{Nado_Arxiv_2020, Schneider_Neurips_2020} & 67.2 & 63.0 & 60.4 & 53.3 & 57.0 & 60.6 & 43.3 & 50.7 & 36.0 & 54.5 & 61.0 & 59.4 & 53.0 & 60.8 & 56.7 & 55.8  \\ 
BN~\cite{Schneider_Neurips_2020} & 89.2 & 21.1 & 68.4 & 79.1 & 68.4 & 76.0 & 60.0 & 46.1 & 26.1 & 81.5 & 72.9 & 76.6 & 55.0 & 81.5 & 75.4 & 65.1  \\ 
TENT~\cite{Wang_ICLR_2021} & 67.2 & 63.0 & 60.4 & 53.3 & 57.1 & 60.6 & 43.3 & 50.8 & 36.1 & 54.6 & 61.1 & 59.5 & 53.2 & 60.9 & 56.6 & 55.8  \\ 
AugBN & 88.2 & 57.3 & 80.4 & 76.1 & 77.1 & 80.0 & 60.9 & 57.7 & 40.0 & 76.4 & 81.2 & 77.5 & 62.7 & 81.5 & 80.6 & \textbf{71.8}  \\ 
	\bottomrule
	\end{tabular}
	}
\end{table*}

\begin{table*} [!htbp]
	\caption{Results for ImageNet-C dataset, with severity level 1 corruptions.
	}
% 	\vspace{1mm}
	\label{table:imagenet_sev1}
% 	\scriptsize
	\centering
	\renewcommand{\arraystretch}{1.7}
	\resizebox{\textwidth}{!}{
	\begin{tabular}{lccccccccccccccccc}
		\toprule
		
		%& & \multicolumn{20}{c}{GTA5 $\rightarrow$ Cityscapes} \\
		%\midrule
		
Method & brightness & contrast & defocus blur & elastic transform & fog & frost & glass blur & gaussian noise & impulse noise & jpeg compression & motion blur & pixelate & shot noise & snow & zoom blur & mCA  \\ 
 \midrule 
Source & 73.9 & 66.0 & 58.2 & 68.2 & 65.5 & 62.2 & 54.8 & 65.0 & 55.6 & 65.5 & 64.6 & 58.8 & 63.6 & 55.5 & 52.2 & 62.0  \\ 
PTN~\cite{Nado_Arxiv_2020, Schneider_Neurips_2020} & 1.0 & 0.9 & 0.4 & 0.7 & 0.8 & 0.7 & 0.5 & 0.7 & 0.6 & 0.7 & 0.8 & 0.6 & 0.7 & 0.6 & 0.5 & 0.7  \\ 
BN~\cite{Schneider_Neurips_2020} & 73.6 & 68.0 & 60.6 & 69.0 & 66.4 & 63.0 & 58.1 & 64.5 & 57.2 & 65.3 & 66.7 & 61.3 & 62.9 & 59.6 & 54.1 & 63.4 \\ 
TENT~\cite{Wang_ICLR_2021} & 0.9 & 0.9 & 0.4 & 0.7 & 0.7 & 0.7 & 0.5 & 0.7 & 0.6 & 0.7 & 0.8 & 0.6 & 0.8 & 0.6 & 0.5 & 0.7  \\ 
% AugBN & 73.5 & 67.7 & 60.2 & 68.6 & 66.0 & 62.5 & 57.4 & 64.8 & 56.8 & 64.8 & 66.1 & 60.5 & 62.9 & 58.9 & 53.6 & 63.0  \\ 
AugBN & 73.9 & 68.2 & 60.8 & 69.3 & 66.6 & 63.4 & 57.9 & 65.0 & 57.5 & 65.8 & 66.7 & 61.1 & 63.4 & 59.9 & 54.0 & \textbf{63.6}  \\ 

	\bottomrule
	\end{tabular}
	}
\end{table*}

\begin{table*} [!htbp]
	\caption{Results for ImageNet-C dataset, with severity level 2 corruptions.
	}
% 	\vspace{1mm}
	\label{table:imagenet_sev2}
% 	\scriptsize
	\centering
	\renewcommand{\arraystretch}{1.7}
	\resizebox{\textwidth}{!}{
	\begin{tabular}{lccccccccccccccccc}
		\toprule
		
		%& & \multicolumn{20}{c}{GTA5 $\rightarrow$ Cityscapes} \\
		%\midrule
		
Method & brightness & contrast & defocus blur & elastic transform & fog & frost & glass blur & gaussian noise & impulse noise & jpeg compression & motion blur & pixelate & shot noise & snow & zoom blur & mCA  \\ 
 \midrule 
Source & 72.2 & 60.1 & 50.6 & 47.5 & 61.3 & 46.6 & 41.1 & 56.4 & 46.8 & 62.3 & 49.7 & 62.0 & 52.9 & 32.3 & 41.8 & 52.2  \\ 
PTN~\cite{Nado_Arxiv_2020, Schneider_Neurips_2020} & 0.9 & 0.8 & 0.3 & 0.4 & 0.7 & 0.5 & 0.3 & 0.6 & 0.5 & 0.6 & 0.5 & 0.7 & 0.6 & 0.4 & 0.4 & 0.5  \\ 
BN~\cite{Schneider_Neurips_2020} & 72.0 & 63.2 & 53.3 & 49.1 & 63.3 & 49.5 & 45.6 & 55.8 & 47.9 & 61.8 & 56.5 & 63.6 & 52.0 & 41.9 & 44.1 & 54.6  \\ 
TENT~\cite{Wang_ICLR_2021} & 0.9 & 0.8 & 0.4 & 0.4 & 0.7 & 0.5 & 0.3 & 0.6 & 0.5 & 0.6 & 0.5 & 0.6 & 0.6 & 0.4 & 0.4 & 0.5  \\ 
% AugBN & 72.0 & 63.2 & 53.4 & 48.9 & 63.2 & 49.4 & 45.4 & 56.0 & 48.0 & 61.5 & 56.1 & 63.3 & 52.1 & 41.6 & 44.2 & \textbf{54.6}\\ 
AugBN & 72.3 & 63.6 & 53.6 & 49.4 & 63.4 & 49.8 & 45.6 & 56.4 & 48.3 & 62.3 & 55.8 & 63.5 & 52.5 & 41.3 & 44.0 & \textbf{54.8}  \\ 

	\bottomrule
	\end{tabular}
	}
\end{table*}

\begin{table*} [!htbp]
	\caption{Results for ImageNet-C dataset, with severity level 3 corruptions.
	}
% 	\vspace{1mm}
	\label{table:imagenet_sev3}
% 	\scriptsize
	\centering
	\renewcommand{\arraystretch}{1.7}
	\resizebox{\textwidth}{!}{
	\begin{tabular}{lccccccccccccccccc}
		\toprule
Method & brightness & contrast & defocus blur & elastic transform & fog & frost & glass blur & gaussian noise & impulse noise & jpeg compression & motion blur & pixelate & shot noise & snow & zoom blur & mCA  \\ 
 \midrule 
Source & 69.3 & 47.8 & 36.4 & 55.4 & 55.8 & 35.2 & 17.8 & 41.3 & 38.5 & 60.0 & 28.2 & 47.7 & 38.7 & 36.0 & 35.4 & 42.9  \\ 
PTN~\cite{Nado_Arxiv_2020, Schneider_Neurips_2020} & 0.8 & 0.7 & 0.3 & 0.5 & 0.6 & 0.3 & 0.2 & 0.5 & 0.4 & 0.5 & 0.4 & 0.5 & 0.5 & 0.4 & 0.3 & 0.5  \\ 
BN~\cite{Schneider_Neurips_2020} & 69.2 & 45.8 & 35.9 & 55.4 & 55.2 & 34.2 & 17.6 & 41.4 & 38.6 & 59.8 & 29.1 & 47.6 & 38.9 & 36.4 & 35.5 & 42.7  \\ 
TENT~\cite{Wang_ICLR_2021} & 0.8 & 0.7 & 0.3 & 0.5 & 0.6 & 0.3 & 0.2 & 0.5 & 0.4 & 0.5 & 0.4 & 0.5 & 0.5 & 0.4 & 0.3 & 0.5  \\ 
% AugBN & 67.2 & 55.6 & 36.9 & 58.0 & 57.9 & 40.2 & 24.4 & 39.6 & 39.0 & 56.6 & 39.6 & 48.4 & 37.9 & 45.1 & 36.8 &\textbf{ 45.5 } \\ 
AugBN & 69.7 & 54.3 & 39.6 & 59.0 & 58.7 & 39.7 & 22.7 & 41.7 & 39.9 & 59.6 & 38.2 & 49.9 & 39.3 & 44.5 & 37.8 & \textbf{46.3}  \\ 

	\bottomrule
	\end{tabular}
	}
\end{table*}

\begin{table*} [!htbp]
	\caption{Results for ImageNet-C dataset, with severity level 4 corruptions.
	}
% 	\vspace{1mm}
	\label{table:imagenet_sev4}
% 	\scriptsize
	\centering
	\renewcommand{\arraystretch}{1.7}
	\resizebox{\textwidth}{!}{
	\begin{tabular}{lccccccccccccccccc}
		\toprule
		
Method & brightness & contrast & defocus blur & elastic transform & fog & frost & glass blur & gaussian noise & impulse noise & jpeg compression & motion blur & pixelate & shot noise & snow & zoom blur & mCA  \\ 
 \midrule 
Source & 64.9 & 22.7 & 25.4 & 41.1 & 53.4 & 33.0 & 12.7 & 23.0 & 20.2 & 51.7 & 12.4 & 31.1 & 18.4 & 25.1 & 28.6 & 30.9  \\ 
PTN~\cite{Nado_Arxiv_2020, Schneider_Neurips_2020} & 0.7 & 0.4 & 0.2 & 0.4 & 0.6 & 0.4 & 0.2 & 0.3 & 0.3 & 0.4 & 0.3 & 0.3 & 0.3 & 0.3 & 0.3 & 0.4  \\ 
BN~\cite{Schneider_Neurips_2020} & 65.3 & 30.8 & 27.7 & 47.6 & 56.0 & 37.6 & 17.0 & 24.4 & 22.7 & 50.4 & 23.2 & 34.1 & 20.4 & 34.7 & 31.2 & 34.9  \\ 
TENT~\cite{Wang_ICLR_2021} & 0.7 & 0.4 & 0.2 & 0.4 & 0.6 & 0.4 & 0.2 & 0.3 & 0.3 & 0.4 & 0.2 & 0.4 & 0.3 & 0.3 & 0.3 & 0.4  \\ 
% AugBN & 63.5 & 35.8 & 26.0 & 48.6 & 55.2 & 38.7 & 18.3 & 24.3 & 23.4 & 47.4 & 24.5 & 32.8 & 20.8 & 35.5 & 30.5 & \textbf{35.0  }\\ 
AugBN & 66.0 & 31.5 & 28.2 & 47.5 & 56.3 & 37.7 & 16.7 & 24.7 & 22.7 & 51.3 & 21.7 & 34.6 & 20.4 & 34.2 & 30.9 & \textbf{35.0 } \\

	\bottomrule
	\end{tabular}
	}
\end{table*}

\begin{table*} [!htbp]
	\caption{Results for ImageNet-C dataset, with severity level 5 corruptions.
	}
% 	\vspace{1mm}
	\label{table:imagenet_sev5}
% 	\scriptsize
	\centering
	\renewcommand{\arraystretch}{1.7}
	\resizebox{\textwidth}{!}{
	\begin{tabular}{lccccccccccccccccc}
		\toprule
		
		%& & \multicolumn{20}{c}{GTA5 $\rightarrow$ Cityscapes} \\
		%\midrule
Method & brightness & contrast & defocus blur & elastic transform & fog & frost & glass blur & gaussian noise & impulse noise & jpeg compression & motion blur & pixelate & shot noise & snow & zoom blur & mCA  \\ 
 \midrule 
Source & 58.2 & 5.8 & 17.2 & 15.5 & 42.9 & 26.4 & 8.7 & 7.3 & 7.3 & 39.7 & 7.2 & 20.8 & 8.8 & 18.8 & 22.8 & 20.5  \\ 
PTN~\cite{Nado_Arxiv_2020, Schneider_Neurips_2020} & 0.6 & 0.2 & 0.2 & 0.2 & 0.5 & 0.3 & 0.1 & 0.2 & 0.2 & 0.2 & 0.2 & 0.3 & 0.3 & 0.3 & 0.2 & 0.3  \\ 
BN~\cite{Schneider_Neurips_2020} & 59.5 & 10.8 & 19.1 & 24.7 & 46.7 & 31.5 & 11.9 & 9.3 & 9.9 & 38.9 & 15.5 & 24.8 & 11.5 & 28.2 & 25.0 & 24.5  \\ 
TENT~\cite{Wang_ICLR_2021} & 0.6 & 0.3 & 0.2 & 0.2 & 0.4 & 0.3 & 0.1 & 0.2 & 0.2 & 0.3 & 0.2 & 0.3 & 0.3 & 0.3 & 0.2 & 0.3  \\ 
AugBN & 58.1 & 13.8 & 17.8 & 28.2 & 46.8 & 32.9 & 12.6 & 10.4 & 11.1 & 35.4 & 17.6 & 23.3 & 12.4 & 30.1 & 24.9 & \textbf{25.0}  \\ 
	\bottomrule
	\end{tabular}
	}
\end{table*}

\end{appendices}

\end{document}